\documentclass[10pt,twocolumn,letterpaper]{article}

 \usepackage[pagenumbers]{cvpr} % To force page numbers, e.g. for an arXiv version
\usepackage{booktabs}
\usepackage{multirow}
\usepackage[table]{xcolor}
\definecolor{beierblue}{RGB}{0, 102, 204}
\definecolor{lightred}{RGB}{240,128,128}

\usepackage{pifont}% http://ctan.org/pkg/pifont
\newcommand{\cmark}{\ding{51}}%
\newcommand{\xmark}{\ding{55}}%

\usepackage{graphicx}
\usepackage{subcaption}
\usepackage[ruled,vlined,linesnumbered]{algorithm2e} % options optional
\usepackage{amsmath,amssymb}
\definecolor{cvprblue}{rgb}{0.21,0.49,0.74}
\usepackage[pagebackref,breaklinks,colorlinks,allcolors=cvprblue]{hyperref}
\usepackage{tikz}
\usetikzlibrary{arrows.meta,positioning,fit,calc,shapes.symbols,shapes.multipart}
\def\paperID{21206} % *** Enter the Paper ID here
\def\confName{CVPR}
\def\confYear{2026}

\title{PAS : Prelim Attention Score for Detecting Object Hallucinations in Large Vision--Language Models}

\author{
Nhat Hoang-Xuan\textsuperscript{1,2} \quad
Minh Vu\textsuperscript{1} \quad
My T. Thai\textsuperscript{2} \quad
Manish Bhattarai\textsuperscript{1} \\
\textsuperscript{1}Los Alamos National Laboratory, Los Alamos, NM, USA \quad
\textsuperscript{2}University of Florida, Gainesville, FL, USA \\
{\tt\small xhoang@lanl.gov \quad mvu@lanl.gov \quad mythai@cise.ufl.edu \quad ceodspspectrum@lanl.gov}
}

\begin{document}
\maketitle
\begin{abstract}
% Large Vision–Language Models (LVLMs) exhibit strong capabilities but remain unreliable due to object hallucinations.
% A lightweight, training-free object hallucination detector can quickly identify hallucinations during inference, enabling real-time filtering and intervention \mvu{bad}. 
% Most training-free detectors focus on image tokens and their consistency with object tokens, while overlooking the role of preceding output tokens (prelim tokens) \mvu{bad}. 
% In this work, we observe that in many hallucinatory predictions, the LVLM neglects the image and instead heavily relies on the preliminary output tokens to infer new objects. We quantify this behavior via the mutual information between the image and the predicted object conditioned on the prelim, showing that a weak dependence on the image strongly correlates with hallucinations. Based on this finding, we introduce Prelim Attention Score (PAS), a novel score computed from prelim tokens attention weights. PAS is efficient since it requires no extra forward passes of the LVLM. By exploiting previously overlooked signals in the prelim, PAS achieves state-of-the-art object hallucination detection performance, demonstrated across multiple models and datasets.

Large vision–language models (LVLMs) are powerful, yet they remain unreliable due to object hallucinations. In this work, we show that in many hallucinatory predictions the LVLM effectively ignores the image and instead relies on previously generated output (“prelim”) tokens to infer new objects. We quantify this behavior via the mutual information between the image and the predicted object conditioned on the prelim, demonstrating that weak image dependence strongly correlates with hallucination. Building on this finding, we introduce the Prelim Attention Score (PAS), a lightweight, training-free signal computed from attention weights over prelim tokens. PAS requires no additional forward passes and can be computed on the fly during inference. Exploiting this previously overlooked signal, PAS achieves state-of-the-art object-hallucination detection across multiple models and datasets, enabling real-time filtering and intervention.

\end{abstract}    
\section{Introduction}
\label{sec:intro}

Following the success of Large Language Models (LLMs), Large Vision-Language Models (LVLMs) brought visual perception and natural language reasoning together. They achieve new frontiers on tasks including but not limited to spatial reasoning~\cite{chen_spatialvlm_2024}, medical visual question answering~\cite{xu_mlevlm_2024}, and visual document understanding~\cite{li_enhancing_2024}. However, hallucination remains a persistent problem, and multimodal capabilities come with a new type of hallucination: \emph{object hallucination}, in which the model mentions objects that are not present in the image~\cite{rohrbach2018chair}. Thus, detecting object hallucination in LVLMs is crucial for ensuring their credibility and reliability, especially in practical scenarios requiring factual correctness~\cite{sahoo_comprehensive_2024}. 

Many existing approaches to object hallucination detection rely on annotations/references~\cite{rohrbach2018chair,li2023pope} or external judge models~\cite{xiao_detecting_2025,jing_faithscore_2024,yin_woodpecker_2024}. These approaches are limited by the lack of references in many real-world cases and the potential unreliability of judge models~\cite{park2025glsim}. Training-free, reference-free approaches avoid this problem by relying only on \emph{intrinsic} information (inputs/outputs and model internals) for detection. 

Effectively utilizing the limited intrinsic information from the LVLM inference process is vital for training-free, reference-free approaches. These methods have considered various sources, such as output logits~\cite{malinin2021uncertainty,zhou2024analyzing} or image token attention~\cite{jiang2025devils}. Notably, they tend to focus on most informative tokens, namely image tokens, while crucially overlooking the low-information preliminary output tokens (which we call the prelim). In contrast, we argue that the model's dependence on this low-information part of the input can signify unreliability, and the inclusion of this information can help with object hallucination detection. We develop our hypothesis with a novel information-theoretic formulation and support our hypothesis with evidence (\cref{fig:violin}, \cref{tab:induce_score}). From our findings, we propose PAS to quantify the dependence of an object token on the prelim, which effectively acts as an object hallucination detector. PAS is based on attention weights computed during model inference, hence its overhead is minimal. \Cref{fig:mainfig} illustrates our findings and our proposed method. Our primary contributions are presented below:
\begin{itemize}
    \item We investigate and promote the role of \emph{preliminary output tokens} in determining object hallucination, a source of information overlooked by previous methods when constructing a detector.
    \item We theorize that excessive focus on prelim tokens can be linked to an unreliable, more hallucinatory alternative operating mode of the LVLM. We provide an information-theoretic formalization to validate our hypothesis.
    \item Based on the above observations, we propose PAS, an attention-based score that achieves state-of-the-art object hallucination detection performance across models and datasets.
\end{itemize}

\paragraph{Organization.} In \Cref{sec:related}, we review existing object hallucination detection methods for LVLM and related lines of research. \Cref{sec:preliminary} elaborates on LVLM token generation dependencies and formalizes the detection problem. In \Cref{sec:motivation}, we theorize the role of the prelim in object hallucination and develop an information-theoretic method to quantify this for detection. We also propose PAS score, an alternative efficient attention-based approach. \Cref{sec:methodology} details the realization of both approaches. \Cref{sec:experiments} presents experiments, ablations, and analyses. Finally, \Cref{sec:conclusion} concludes and discusses limitations.
\section{Related Work}
\label{sec:related}

\paragraph{Object Hallucination Detection.} Object hallucination is a phenomenon where the model outputs refer to objects that are inconsistent with or do not exist in the input images~\cite{li2023pope,park2025glsim,rohrbach2018chair}. Existing  approaches to object hallucination detection can be classified based on the information that they use:
\begin{itemize}
    \item \textbf{Logits-based approaches}~\cite{malinin2021uncertainty,zhou2024analyzing} primarily rely on just the outputs of the LVLM for hallucination detection. They generally compute some kind of information-theoretic uncertainty measure based on probabilistic interpretation of the token logits. While principled, since they use the least information (compared to other categories), they tend to have worse detection performance.
    \item \textbf{Representation-based approaches}~\cite{jiang_interpreting_2024,park2025glsim} typically focus on image tokens and their compatibility with predicted objects. They leverage Logit Lens~\cite{noauthor_interpreting_2020} to map intermediate image hidden states to output probabilities and use this as a proxy for image-object compatibility. 
    
    % While hidden states is most rich in information, it is also abstract and most difficult to interpret.
    \item \textbf{Attention-based approaches}~\cite{jiang2025devils} leverage the attention weights to quantify the dependency of output tokens on input tokens. However, similar to representation-based approaches, they mostly focus on image tokens.
\end{itemize}

Existing approaches tend to focus on image tokens and overlook information in the prelim tokens. In contrast, we observe that frequently when hallucination occurs, the model exhibits high dependency on prelim tokens. This observation is detailed in \Cref{sec:prelim_matters}. Based on this observation, we develop a detector that takes the prelim into account, and obtain an efficient hallucination detection that achieves state-of-the-art performance.

\paragraph{Related research directions.}
Apart from \textit{internal} information, some object hallucination methods leverage \textit{external} information, such as an external oracle~\cite{xiao_detecting_2025,chen_unified_2024,yin_woodpecker_2024,jing_faithscore_2024}. In many cases, this reliance can be problematic since bigger models are costly to run and the external models themselves can be wrong. In this work, we focus on the setting where no external information is used and no additional training is performed, and compare against baselines with the same assumptions. Furthermore, some existing works focus on \textit{sentence}-level or \textit{segment}-level hallucination
in LVLMs~\cite{li_reference-free_2024,gunjal_detecting_2024,xiao_detecting_2025}, while object hallucination is \textit{token}-level, and thus those approaches cannot be directly applied to this task. Finally, a line of research~\cite{huang2024opera,liu_paying_2024,an2025mitigating,leng2024vcd} focuses on \emph{mitigation} of object hallucination, which directly alters the model, its inputs, or the generation process in order to reduce hallucination. Mitigation methods do not directly lead to detection and also do not eliminate all hallucinations, hence our work and mitigation methods are complementary. We provide additional discussion of related works in the Appendix.

\section{Preliminary}
\label{sec:preliminary}
\begin{figure*}
    \centering
    \includegraphics[clip,trim={0 4cm 0 0},width=\linewidth]{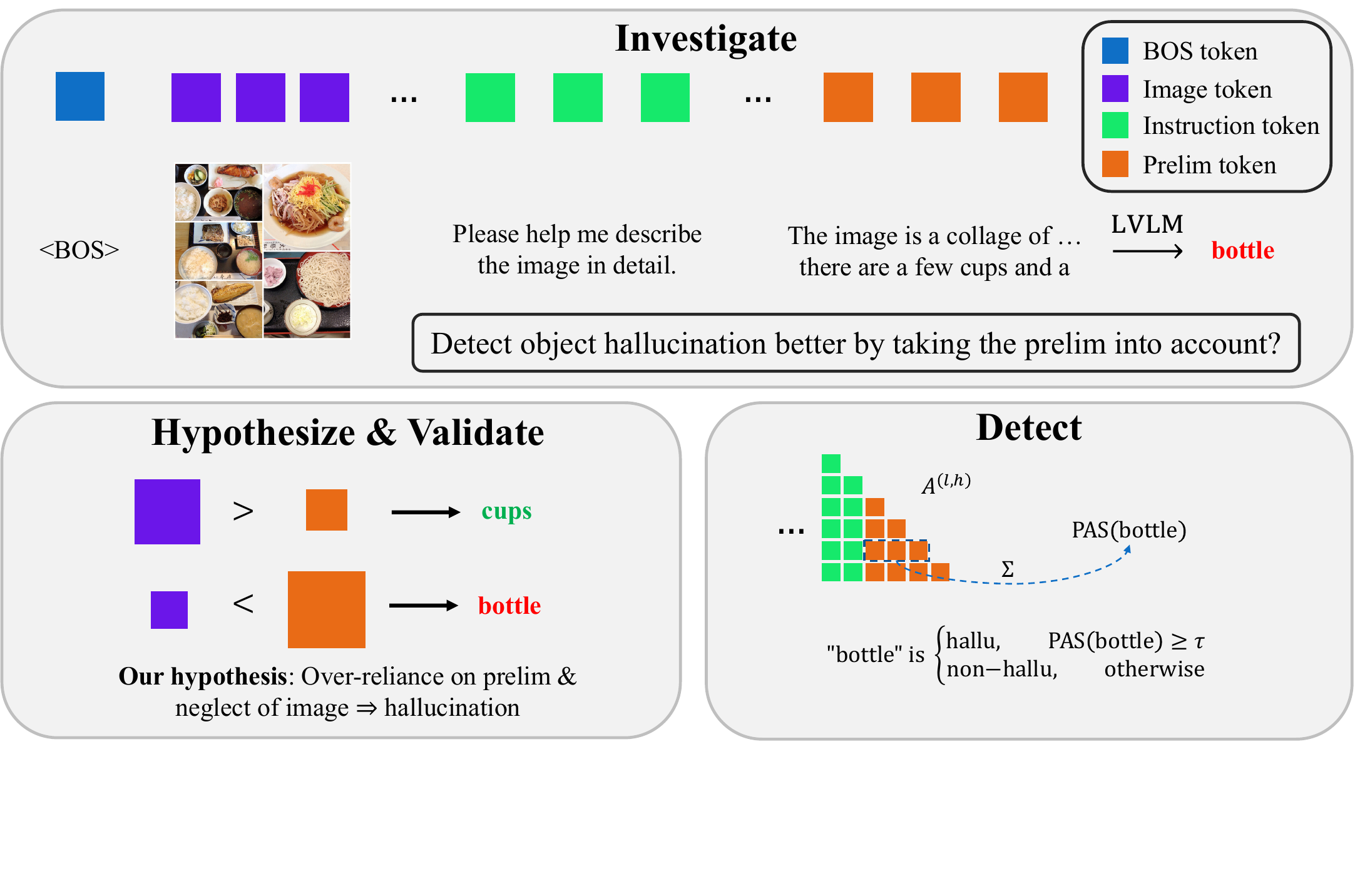}
    \caption{Illustration of our findings and proposed method. \textbf{Top (Investigate)}: We show that LVLM token generation depends on four token types and argue that prelim tokens are a vital, overlooked signal. \textbf{Bottom Left (Hypothesize \& Validate)}: We visually illustrate our core hypothesis: hallucinations (e.g., ``bottle'') occur when the model over-relies on prelim tokens, while real objects (e.g., ``cups'') rely more on the image. \textbf{Bottom Right (Detect)}: Based on this insight, we propose PAS, which quantifies this over-reliance by summing the attention weights from prelim tokens to the object token.}
    \label{fig:mainfig}
\end{figure*}

\paragraph{Notation.} Let $\Phi$ be the LVLM.
% , which takes in a sequence of tokens and output a probability distribution $p \in \mathbb{R}^V$ over the vocabulary $V$
Let $\mathbf{v},\mathbf{t}$ denote the image tokens and instruction tokens, respectively. Let $\mathbf{x} = (\mathbf{v},\mathbf{t})$ be the input tokens, and $\mathbf{y}$ be the output tokens. Note that we omit the BOS token in $\mathbf{x}$ for brevity. Let $m = |\mathbf{x}|$ be the number of input tokens, and $n = |\mathbf{x}| + |\mathbf{y}|$ be the total number of tokens. 

% In other words, the $k$-th output token $\mathbf{y}_k$ only depends on the input $\mathbf{x}$ and previous output tokens $\mathbf{y}_{<k}$. 

\paragraph{Token generation dependency.} In this work, we focus on LVLMs which are autoregressive decoders. For autoregressive LVLMs, the probability of producing the output sequence $\mathbf{y}$ is given as:

\begin{equation}
\label{eq:autoregressive}
    \Pr(\mathbf{y} \mid \mathbf{x}) = \prod_{k=m+1}^n \Pr(y_k \mid \mathbf{y}_{< k}, \mathbf{x})
\end{equation}

\noindent
where 

\begin{equation}
\label{eq:token_dist}
    \Pr(\cdot \mid \mathbf{y}_{< k}, \mathbf{x}) = \operatorname*{softmax}\left( \Phi(\cdot \mid \mathbf{y}_{<k}, \mathbf{x}) \right)
\end{equation}

\noindent
represents the output of a single forward pass of the model $\Phi$.
From ~\cref{eq:autoregressive}, the generation of $y_k$ depends on the input $\mathbf{x} = (\mathbf{v},\mathbf{t})$ and \emph{previously} generated tokens $\mathbf{y}_{< k}$ (which we call the \emph{prelim}), therefore investigating these components can help us understand hallucinatory predictions. 

As discussed in \Cref{sec:related}, existing approaches tend to extract signals from the image tokens $\mathbf{v}$ since it contains the ``factual'' information that the model should focus on, while the prelim contains incomplete information~\cite{huang2024opera}. Conversely, in this work, we show that strong signals for object hallucination detection can be extracted from the prelim.

\paragraph{Attention mechanism.} The attention mechanism can be used to quantify the dependency of the generation of a token on previous tokens. This is because many popular LVLMs use Large Language Model (LLM) backbones that have decoder-only Transformer architecture (e.g, LLaMA~\cite{touvron_llama_2023}, QwenLM~\cite{bai_qwen_2023}). In this architecture, an LLM consists of only self-attention and MLP layers. Since the MLP layers operate independently on each token, the attention layers are the only path for information exchange between tokens in a single forward pass. Therefore, looking at attention weights can give insight on the information flow. In particular, looking at the attention weights of an output token can reveal which previous tokens significantly affect its generation. In this work, we leverage the attention weights to quantify how much prelim tokens affect generation of object tokens and link it to hallucination.

\paragraph{Object hallucination detector.} When prompting the LVLM with an image $\mathbf{v}$, the output tokens $\mathbf{y}$ can contain references to objects, and we wish to determine if they are present or absence from the image $\mathbf{v}$ (corr. to real or hallucinatory). We define an \emph{object hallucination scoring function} $D(y_k,\mathbf{y},\mathbf{x})$ as a function that takes in an object mention $y_k \in \mathbf{y}$ and returns a real-valued score, where \emph{higher} values indicate a \emph{higher} chance of $y_k$ being a hallucination. Given this function, a detector can be obtained via thresholding with some threshold $\tau$, i.e., all $y_k$ with $D(y_k,\mathbf{y},\mathbf{x}) \geq \tau$ is considered a hallucination.
\section{Detecting Object Hallucination via Prelim}
\label{sec:motivation}

% \mvu{This section describes how the VLM generate tokens based on context, how the prelim are feed as context and our hypothesis on why (our hypothesis 1 (\textbf{H1})) it creates hallucination. Please think about how you want to state the hypothesis here.}

% \mvu{After the hypothesis, establish your approach to validate that hypothesis. I think Figure 3 is a good validation. This is also the place for you to talk about the entropy/KL/logit diff approach and some of their results. }

% \mvu{After that, you can say the drawbacks of those methods, which promotes the attention based method, which we emphasize in the next section}

% overall strategy: move the motivation (the why) parts from Sec Methodology here, and let the method sec focus on the how

% state that autoregressive allows us to look at the prediction of single (object) tokens. Then state (a slightly modified version of) the hypothesis: we hyothesize that the model does not focus on the image and focus heavily on prelim. 

% Validate 2 ways: prelim attention and induce scores. Also talk about why attention works here

% drawback (computationally expensive), then promote attention

In this section, we first describe the LVLM token generating process. We then form our hypothesis on how prelim tokens are important in hallucination detection. Finally, we describe how to validate our hypothesis and discuss its implications, which leads to the methodology in \Cref{sec:methodology}.

\subsection{Prelim matters for hallucination detection}
\label{sec:prelim_matters}

As laid out in \Cref{sec:preliminary}, the generation of a single token $y_k$ depends on image tokens $\mathbf{v}$, instruction tokens $\mathbf{t}$, and prelim tokens $\mathbf{y}_{<k}$. Crucially, the effect of each token type on the generation of $y_k$ depends on how much the model focus on them~\cite{liu_paying_2024,gong_damro_2024}, which can be quantified via the attention mechanism. Existing approaches~\cite{jiang2025devils,liu_paying_2024,an2025mitigating} argue that in the normal operating mode, the model should mostly focus on the image for object generation, implying \textit{low image attention} $\rightarrow$ \textit{higher} chance of hallucination, and base on this to determine hallucinations. Importantly, they do not consider the prelim tokens for object hallucination detection. At first, it might be intuitive to disregard this source of information as the prelim contains much less information than the image, and is also less reliable because it is generated rather than given. However, in fact, the attention on the prelim is vital because if the model relies on unreliable information to generate tokens, then that behavior is undesired. Thus, we hypothesize that the inclusion of information from the prelim, which previous works missed, can help with hallucination detection. Our question is: ``\textbf{Can we create a better object hallucination detector by taking the prelim into account?}''

\begin{figure}
    \centering
    \includegraphics[width=0.95\linewidth]{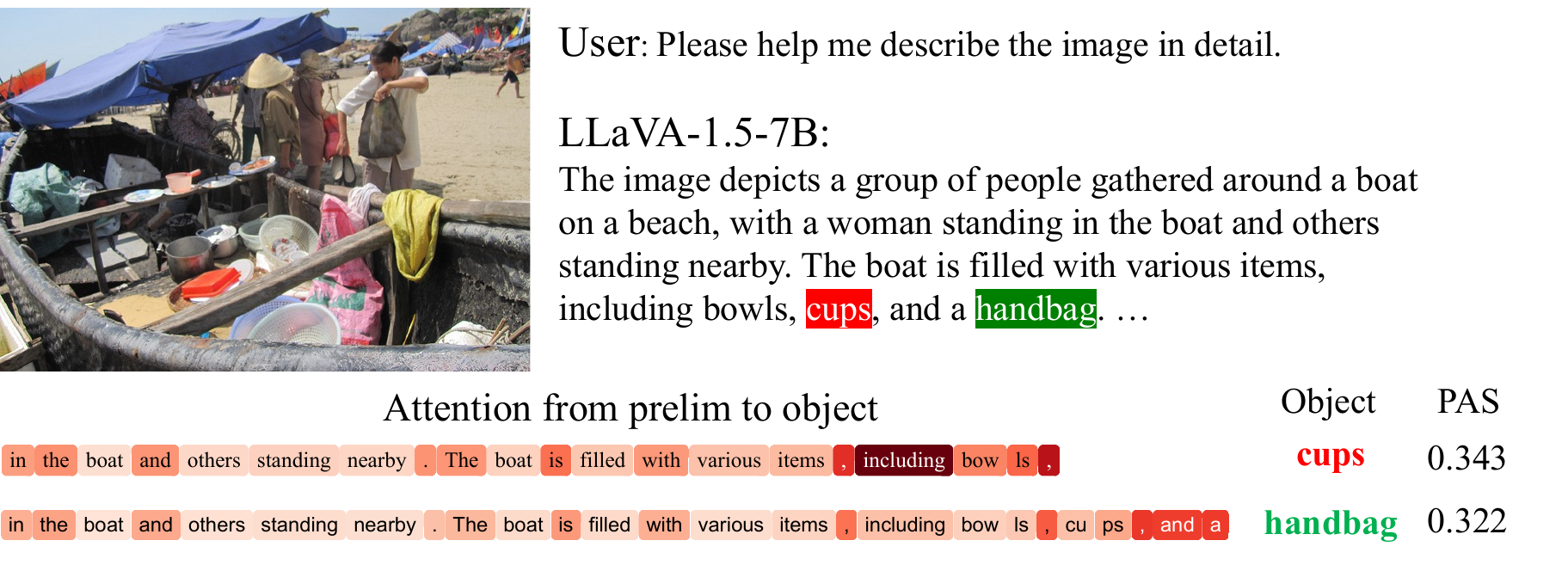}
    \caption{Visualization of Prelim Attention Score (PAS) for a single sample. We show the per-token attention for a suffix of the prelim for a hallucinated and a real object token. Darker red indicates higher attention to the corresponding object token, and a higher PAS score indicates a higher chance of hallucination. }
    \label{fig:qualitative}
\end{figure}

\begin{figure}[t]
    \centering
    \begin{subfigure}[t]{0.48\linewidth}
        \centering
        \includegraphics[width=\linewidth]{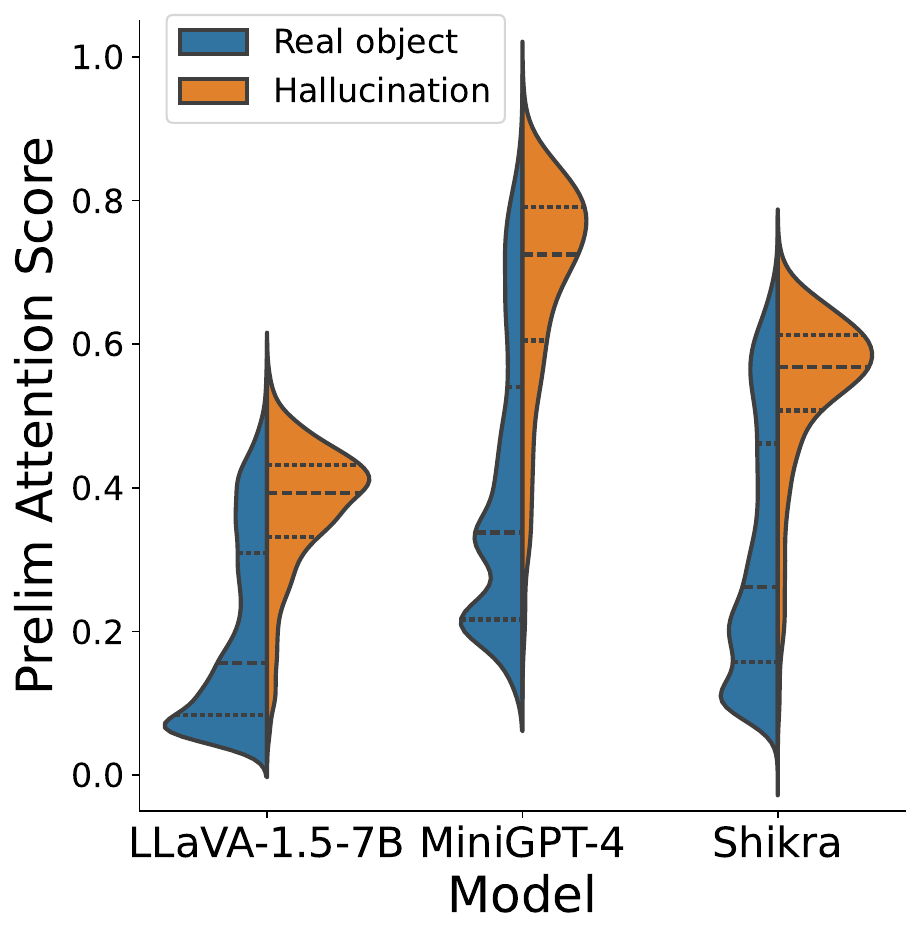}
        \caption{PAS score (\cref{eq:prelim_attention_score})}
        \label{fig:violin_PAS}
    \end{subfigure}%
    ~ 
    \begin{subfigure}[t]{0.48\linewidth}
        \centering
        \includegraphics[width=\linewidth]{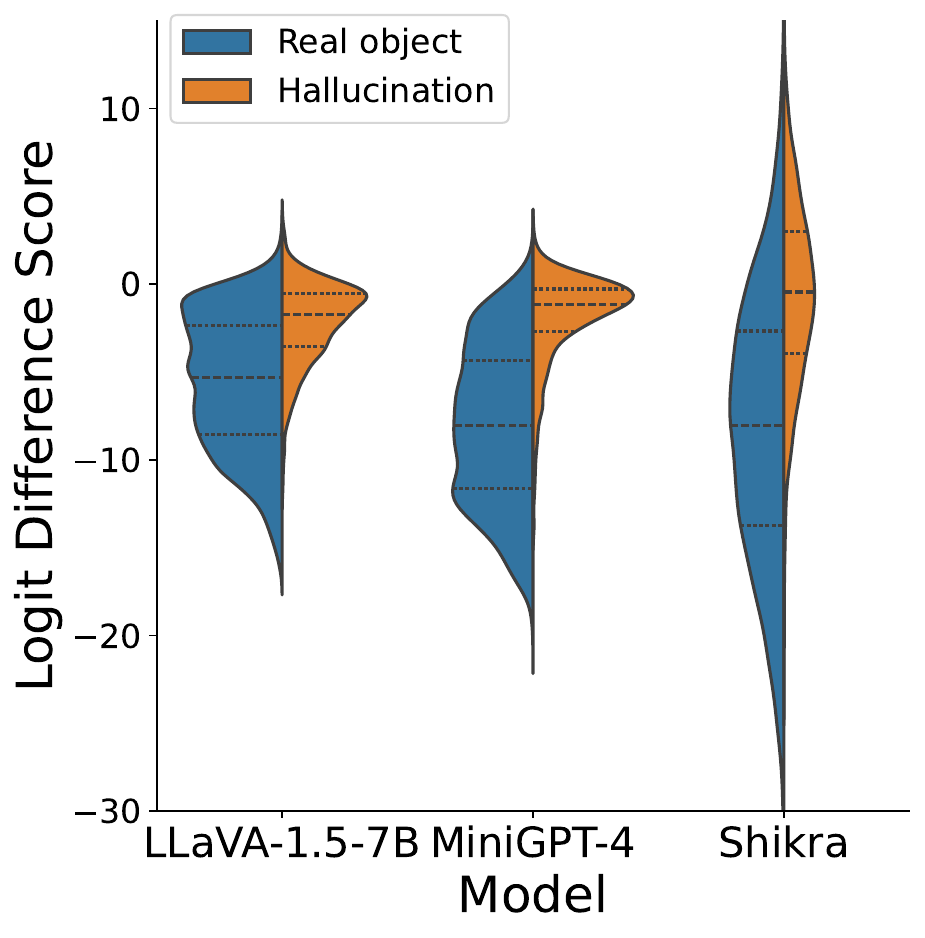}
        \caption{MI-based score (\cref{eq:logit_induce_score})}
        \label{fig:violin_MI}
    \end{subfigure}
    \caption{Score distributions for real and hallucinated object tokens across different models on MSCOCO dataset. The dashed lines denote the quartiles for each distribution.}
    \label{fig:violin}
\end{figure}

In \cref{fig:violin_PAS}, we visualize the sum of attention weights (formally defined in \cref{eq:prelim_attention_score}) from prelim tokens to object tokens, categorized into real and hallucinatory tokens. This suggests that information from prelim tokens can be highly beneficial in determining hallucination. Specifically, the relation is \textit{high prelim attention} $\rightarrow$ \textit{higher} chance of hallucination, which is the inverse of that for image tokens. This raises the question: \textbf{when not attending to the image, why the model focuses on the prelim} instead of ``certainly meaningless tokens'' like the BOS~\cite{barbero_why_2025}?''
% At first, it is not clear why looking at the prelim tokens could help with hallucination, since prelim tokens typically contain either partial, non-informative text (``there is a '') or partial depictions of the image, which is inadequate to confirm the existence of new objects. 
% However, apart from the image tokens themselves, the prelim tokens are still directly derived from the image, unlike the BOS (which is always present), and the instruction which typically only specifies the task.

We hypothesize that the model's focus on the prelim tokens reveals an \emph{alternative operating mode}. In this mode, the model does not base its prediction on the image, possibly due to a lack of sufficient image-perceiving capabilities or because the image itself is noisy and is thus challenging to interpret. Regardless of the true reason, in this mode, the model predominantly depends on the prelim, but since this source of information is imperfect, it leads to more unreliable and hallucinatory predictions. If our hypothesis is true, then recognizing this mode could lead to a more effective hallucination detector, which motivates us to investigate it deeper. In the context of the object hallucination detection problem, we pose the following hypothesis (\textbf{H1}): ``\textbf{If the generation of an object token depends heavily on the prelim tokens and not on the image tokens, then it is likely to be a hallucination.}''

To validate \textbf{H1}, we first define $Y_k$ as the random variable representing the token to be predicted at position $k$. \textbf{H1} says that there are two potential factors that determine $Y_k$: the prelim tokens $\mathbf{y}_{<k}$ and the image tokens $\mathbf{v}$. Moreover, it says that if $Y_k$ is not dependent on $\mathbf{v}$ then $y_k \sim Y_k$ is likely to be a hallucination. This notion can be precisely captured by the notion of mutual information (MI): if $I(\mathbf{v};Y_k \mid \mathbf{y}_{<k},\mathbf{t})$ is small then $y_k$ is likely to be a hallucination. 
% given the prelim $\mathbf{y}_{<k}$, if observing $\mathbf{v}$ yields little additional information about $Y_k$, then $Y_k$ is minimally dependent on $\mathbf{v}$. 
Thus, we propose to use the MI as the detector:

\begin{equation}
\label{eq:detector_MI}
    D_{\text{MI}}(y_k,\mathbf{y},\mathbf{x}) = -I(\mathbf{v};Y_k \mid \mathbf{y}_{<k},\mathbf{t}),
\end{equation}

\noindent
where
\begin{equation}
\label{eq:induce_entropy_diff}
    I(\mathbf{v};Y_k \mid \mathbf{y}_{<k},\mathbf{t}) = H(Y_k \mid \mathbf{y}_{< k}, \mathbf{t}) - H(Y_k\mid \mathbf{y}_{< k}, \mathbf{x}).
\end{equation}

% Intuitively, if the model is in alternative operating mode, then $Y_k$ mostly depends on $\mathbf{y}_{<k}$. In that case, observing $\mathbf{v}$ gives little information about $Y_k$, leading to low mutual information and high $D_\text{MI}$ score. If our hypothesis \textbf{H1} is correct, then this should imply high chance of hallucination. 
The main challenge in computing \cref{eq:induce_entropy_diff} lies in obtaining the distribution of $Y_k \mid \mathbf{y}_{<k}, \mathbf{t}$, which is not directly available from the LVLM. Intuitively, this distribution can be thought of as the ``prediction'' given just the prelim, therefore, to compute $\Pr(Y_k = y \mid \mathbf{y}_{< k},\mathbf{t})$, we propose to marginalize out the image $\mathbf{v}$ from \cref{eq:token_dist}:

\begin{equation}
\label{eq:marginalize_image}
    \Pr(y \mid \mathbf{y}_{< k},\mathbf{t}) = \mathbb{E}_{I \sim \mathcal{I}}\left[ \Pr\left(y \mid \mathbf{y}_{< k},(I, \mathbf{t}) \right) \right].
\end{equation}

We visualize the distribution of \cref{eq:detector_MI} for real and hallucinated tokens in \Cref{fig:violin_MI} and the detection AUROC in \Cref{tab:induce_score}, which support our hypothesis \textbf{H1}. Further details on the implementation is described in \Cref{sec:dependence_prelim}. 
% Finally, in \Cref{sec:results}, we experimentally show that various realizations of this approach lead to an effective object hallucination detector, which supports our hypothesis.

% In other words, $\Pr(Y_k \mid \mathbf{y}_{< k},\mathbf{t})$ is the \textit{average} prediction of the model given the prelim $\mathbf{y}_{< k}$. 
% Given this definition, we propose to use the following score for detecting hallucination:

% Intuitively, by comparing prelim-dependent image-independent probabilities against image-dependent probabilities, we can tell if the image tokens $\mathbf{v}$ played a role in the generation of the token $y_k$. 
% This approach is described in details in \Cref{sec:dependence_prelim}, and in \Cref{sec:results} we experimentally show that this leads to an effective object hallucination detector, which supports our hypothesis.

% If these two distributions are similar, then the prediction is mostly based on the prelim, and we give it a high score indicating high possibility of hallucination. 

\subsection{Prelim Attention-based approach}
\label{sec:attention_approach}

A major issue with using the $D_\text{MI}(y_k,\mathbf{y},\mathbf{x})$ in \cref{eq:detector_MI} for detection is the estimation step~(\ref{eq:marginalize_image}). Averaging over some set of images $\mathcal{I}$ with size $|\mathcal{I}| = L$ necessitates $L$ extra forward passes of the model for each input $\mathbf{x}$. Thus, this approach demands $L + 1$ times more computation than the generation phase, which makes it highly impractical. 

To overcome this, we propose an alternative detector based on the attention mechanism. From \cref{fig:violin_PAS}, \textbf{the model assigns significantly higher attention weights to prelim tokens when generating hallucinatory object tokens, compared to real object tokens}. Let $s_\text{BOS}, s_\text{img}, s_\text{ins},s_\text{prel}$ denote the sum of attention weights (formally defined in \cref{eq:prelim_attention_score}) from respective token types to $y_k$. The softmax operation in self-attention layers requires that $s_\text{BOS} + s_\text{img} + s_\text{ins} + s_\text{prel} = 1$. This property, along with the fact that the attention mechanism is the only method of information exchange between tokens in a forward pass, suggests that attention scores can be used to quantify how much each token types affect generation of object tokens. Thus, in context of our hypothesis \textbf{H1}, we propose to directly use $s_\text{prel}$ for hallucination detection:

\begin{equation}
\label{eq:detector_prelim_attention}
    D_{\text{PAS}}(y_k,\mathbf{y},\mathbf{x}) = s_{\text{prel}}(y_k,\mathbf{y},\mathbf{x}).
\end{equation}

In \Cref{sec:prelim_score}, we describe in details how to compute this score, and in \Cref{sec:results}, we show that this leads to an efficient detector.
\section{Methodology}
\label{sec:methodology}

\begin{table*}[h]
    \centering
    \resizebox{.92\linewidth}{!}{
    \begin{tabular}{l c c c c c c | c} 
        \toprule
        \multirow{2}{*}{\textbf{Variant}} & \multicolumn{2}{c}{\textbf{LLaVA-1.5-7B}} & \multicolumn{2}{c}{\textbf{MiniGPT-4}} & \multicolumn{2}{c}{\textbf{Shikra}} & \multirow{2}{*}{\textbf{Average}} \\
        \cmidrule(lr){2-3} \cmidrule(lr){4-5} \cmidrule(lr){6-7}
        & MSCOCO & Pascal VOC & MSCOCO & Pascal VOC & MSCOCO & Pascal VOC & \\
        \midrule
        Entropy diff (\ref{eq:induce_entropy_diff}) & 71.7 & 76.4 & 84.0 & 82.8 & 64.9 & 67.2 & 74.5 \\
        KL div (\ref{eq:kl_induce_score}) & 73.8 & 81.1 & 86.5 & 86.2 & 73.1 & 80.4 & 80.2 \\
        Logit diff (\ref{eq:logit_induce_score}) & 76.6 & 81.9 & \textbf{88.1} & \textbf{87.4} & 79.0 & 80.3 & 82.2 \\
        PAS (\ref{eq:detector_prelim_attention})  & \textbf{84.2} & \textbf{85.1} & 85.6 & 85.4 & \textbf{84.5} & \textbf{85.3} & \textbf{85.0} \\
        \bottomrule
    \end{tabular}
    }
    \caption{Comparing detection performance (AUROC, higher is better) of different choices of $\Delta$ in \cref{eq:induce_score} against attention-based PAS across different models and datasets. All values are percentages, and best results are shown in \textbf{bold}.}
    \label{tab:induce_score}
\end{table*}

\subsection{Mutual Information-based score}
\label{sec:dependence_prelim}

Since the estimation in \cref{eq:marginalize_image} can introduce some errors, we also consider other candidates for computing $I(\mathbf{v}; Y_k \mid \mathbf{y}_{<k}, \mathbf{t})$, apart from the difference in entropy in \cref{eq:induce_entropy_diff}. The detector in \cref{eq:detector_MI} can be generalized as follows:

\begin{equation}
\label{eq:induce_score}
    D_\Delta(y_k, \mathbf{y}, \mathbf{x}) = \Delta\left( Y_k \mid \mathbf{y}_{< k},\mathbf{t}; Y_k \mid \mathbf{y}_{<k}, \mathbf{x} \right)
\end{equation}

\noindent
in which $\Delta$ is a function that computes some ``distance'' between two discrete probability distributions. To realize \cref{eq:induce_score}, there are two points of focus: the estimation in \cref{eq:marginalize_image}, and the specification of $\Delta$.

\paragraph{Estimation of $\Pr(y \mid \mathbf{y}_{< k},\mathbf{t})$.} Since it is impossible to sample from the (unknown) distribution of all images, we estimate by averaging over a fixed set of images $\mathcal{I}$ instead. We lay out some strategies for a more stable estimation. We sample once per dataset, letting $\mathcal{I}$ be a subset of that dataset and use the same $\mathcal{I}$ for different prelims. Furthermore, since $y$ varies over the set of object tokens, ideally, we want $\mathcal{I}$ to contain all possible objects so that the model's true prediction tendency given a real object is observed for all object classes. This is achieved by randomly sampling an image that contains each class for all object classes. Thus, we have $|\mathcal{I}| = $ number of object classes in the dataset.

\paragraph{Choice of $\Delta$.} First, we note that the entropy of a distribution is frequently difficult to compute. In our case, we can compute the entropy because the next token has only finite possibilities. Apart from the entropy difference as in (\ref{eq:induce_entropy_diff}), we consider two other candidates for quantifying the ``distance'' between the two discrete probability distributions. The first one is the KL Divergence, which is commonly used in the literature:

\begin{equation}
\label{eq:kl_induce_score}
    D_{\text{div}}(y_k, \mathbf{y}, \mathbf{x}) = -D_\text{KL}\left( Y_k \mid \mathbf{y}_{< k},\mathbf{t} \middle|| Y_k \mid \mathbf{y}_{<k}, \mathbf{x} \right).
\end{equation}

% Thus, there are many candidates for $\Delta$ from probability theory and information theory. 
Alternatively, we directly look at the difference in the predicted logit (pre-softmax) of the object token $y_k$. Intuitively, the logit can be interpreted as (the inverse of) the uncertainty for a prediction, which gives the following score: 

\begin{multline}
\label{eq:logit_induce_score}
    D_{\text{logit}}(y_k, \mathbf{y}, \mathbf{x}) = \left( \sum_{I \in \mathcal{I}} \frac{\Phi(y_k \mid \mathbf{y}_{< k},(I, \mathbf{t}))}{|\mathcal{I}|} \right) \\ - \Phi(y_k \mid \mathbf{y}_{<k}, \mathbf{x}).
\end{multline}

% Thus, we expect the model to mainly focus on image tokens when generating object tokens. However, when studying hallucination samples, we observe that the model tends to focus on preliminary tokens when generating hallucinatory tokens. While the preliminary output can contain certain information regarding the image content, it cannot be used to predict with absolute certainty whether an unmentioned object is present or not. 

% Intuitively, the score $f(y_k,\mathbf{y},\mathbf{x})$ quantifies the following: given the prelim $\mathbf{y}_{<k}$, how much does the image $\mathbf{v}$ change the prediction of the model? If the score is low, it suggests that the model tend to predict the same object probabilities regardless of the presence of the image, which suggests that it relies heavily on the prelim, making it prone to hallucination. To quantify the difference between the two probability distributions $\Pr(y_k \mid \mathbf{y}_{< k},\mathbf{t})$ and $\Pr(y_k \mid \mathbf{y}_{<k}, \mathbf{x})$, we experimented with different $\Delta$ functions: entropy difference $\Delta(p, q) = H(q) - H(p)$, KL divergence $\Delta(p, q) = -D_{KL}\left(p \middle\| q\right)$, and finally token logit difference $\Delta(p, q) = p(y_k) - q(y_k)$.

As shown in \Cref{tab:induce_score}, the variants lead to functional detectors with varying degrees of effectiveness.

% However, the estimation of $\Pr(y_k \mid \mathbf{y}_{< k},\mathbf{t})$ requires a forward pass of the model for each image in $\mathcal{I}$, which makes it prohibitively expensive in practice. 

\subsection{Prelim Attention Score}
\label{sec:prelim_score}

In this section, we describe in details the attention-based approach in \Cref{sec:attention_approach}. We start with some notations. Let $H$ be the number of attention heads in a layer of the LVLM. Let the attention weights at head $h$ of layer $l$ for the sequence $(\mathbf{x}, \mathbf{y})$ be $\mathbf{A}^{(l,h)} \in \mathbb{R}^{n \times n}$. Let $\mathbf{A}^{(l,h)}(k,j)$ denotes the attention from the $j$-th token in $(\mathbf{x},\mathbf{y})$ to object token $y_k$. Then the prelim attention score (using layer $l$) is:

\begin{equation}
\label{eq:prelim_attention_score}
    s_\text{prel}(y_k, \mathbf{y}, \mathbf{x}) = \frac1H \sum_{h=1}^H \sum_{j=m+1}^{k-1} \mathbf{A}^{(l,h)}(k,j).
\end{equation}

% Given the phenomenon in \Cref{sec:dependence_prelim}, we seek to find a different way to quantify how much the model pay attention to the prelim, in comparison to the image. The attention mechanism, which is how decoder-only Transformers transmit information between tokens, suggests an alternative for this.

Different layers in the LVLMs might correspond to different stages of processing of the model~\cite{artzy_attend_2024,gupta_how_2025}, which can lead to varying detection performance. Based on our ablation in \cref{fig:ablate_layer} (discussed further in \cref{sec:ablation_study}), we use the first layer (layer 0) as the default.
Regarding multi-head attention, we average over all heads in a layer for simplicity, following~\cite{jiang2025devils}. While some work~\cite{sarkar_mitigating_2025} suggests that selecting a certain subset of heads can be a better alternative than simply averaging over all heads, this adds additional complexity and requires validation so we leave it for future work. Furthermore, while this work focuses on the prelim, we also experiment with other scores for comparison. In particular, the scores $s_\text{BOS}, s_\text{img}, s_\text{ins}$ can be defined similarly to \cref{eq:prelim_attention_score}, summing over the respective token types, and we experiment with them in \Cref{sec:ablation_study}.

As discussed earlier (\cref{sec:attention_approach}), this attention-based approach is much more computationally efficient than the approach in \Cref{sec:dependence_prelim} since it does not require extra passes and only relies on the attention weights, which are commonly computed during inference. Thus, in subsequent comparisons, unless otherwise stated, we use the prelim attention score of the first layer as our default method.

% For detection purposes, only $s_\text{prel}$ has a positive correlation with hallucination likelihood, while other scores have negative correlation. These scores have significant correlation as depicted in \cref{fig:attn_cor}. Regardless, our ablation study in \Cref{sec:ablation_study} shows that the prelim attention is overall a more reliable signal for hallucination detection.

% Experimentally, we find that higher $s_{prel}$ suggests hallucination, while for the remaining scores, lower values indicate hallucination. The softmax function constrains their sum to 1, so an increase in $s_{prel}$ necessitates a decrease in some of the remaining scores. In \cref{fig:attn_cor}, we note that there is a significant correlation between different attention scores, making them also useful in detecting hallucination (demonstrated in \Cref{tab:ablate_attention_types}). However, this correlation is not consistent across all models, and the prelim attention score remains most effective overall.

% This score directly uses the model's average attention weights to the prelim in a single layer to quantify hallucination. 

%\input{supp/algo}
\section{Experiments}
\label{sec:experiments}

\begin{table*}[t]
    \centering
    \resizebox{.92\linewidth}{!}{
    \begin{tabular}{l c c c c c c | c} 
        \toprule
        \multirow{2}{*}{\textbf{Method}} & \multicolumn{2}{c}{\textbf{LLaVA-1.5-7B}} & \multicolumn{2}{c}{\textbf{MiniGPT-4}} & \multicolumn{2}{c}{\textbf{Shikra}} & \multirow{2}{*}{\textbf{Average}} \\
        \cmidrule(lr){2-3} \cmidrule(lr){4-5} \cmidrule(lr){6-7}
        & MSCOCO & Pascal VOC & MSCOCO & Pascal VOC & MSCOCO & Pascal VOC & \\
        \midrule
        % IC & XX.X & XX.X & XX.X & XX.X & XX.X & XX.X & XX.X \\
        % IC & XX.X & XX.X & XX.X & XX.X & XX.X & XX.X & XX.X \\
        NLL \cite{zhou2024analyzing}    & 56.5 & 64.0 & 62.1 & 73.0 & 54.3 & 63.1 & 62.2 \\
        Entropy \cite{malinin2021uncertainty} & 71.7 & 64.3 & 69.8 & 62.9 & 71.4 & 64.4 & 67.4 \\
        IC \cite{jiang_interpreting_2024}     & 75.1 & 64.6 & 76.4 & 67.7 & 76.0 & 71.3 & 71.9 \\
        GLSim \cite{park2025glsim}   & 64.1 & 69.4 & 63.6 & 62.0 & 67.8 & 66.6 & 65.6 \\
        SVAR \cite{jiang2025devils}    & 81.5 & 82.9 & \textbf{88.0} & 84.5 & 71.9 & 72.9 & 80.3 \\
        Ours    & \textbf{84.2} & \textbf{85.1} & 85.6 & \textbf{85.4} & \textbf{84.5} & \textbf{85.3} & \textbf{85.0} \\
        \bottomrule
    \end{tabular}
    }
    \caption{Object hallucination detection performance (AUROC, higher is better) on three models and across two datasets. For all models, greedy decoding with \texttt{max\_new\_tokens=512} is used. All values are percentages, and best results are shown in \textbf{bold}.}
    \label{tab:main_results}
\end{table*}

\subsection{Setup}

\paragraph{Models.} We conduct experiments on three popular LVLMs: LLaVA-1.5~\cite{liu_visual_2023}, MiniGPT-4~\cite{zhu2024minigpt}, and Shikra~\cite{chen_shikra_2023}. All models have size 7B. For all models, greedy decoding with \texttt{max\_new\_tokens=512} is used. Additional results for bigger models and more implementation details are provided in the Appendix.

\begin{table}[t]
    \centering
    \resizebox{.65\linewidth}{!}{
    \begin{tabular}{l c c c}
        \toprule
        \textbf{Method} & A & H & \textbf{VRAM (GB)} \\
        \midrule
        Entropy & \xmark & \xmark & 16 \\
        IC      & \xmark & \cmark & 30 \\
        GLSim   & \xmark & \cmark & 19 \\
        SVAR    & \cmark & \xmark & 18 \\
        Ours    & \cmark & \xmark & 18 \\
        \bottomrule
    \end{tabular}
    }
    \caption{Memory consumption of each detection method (\textbf{batch size 1}) when using LLaVA-1.5-7B in \texttt{float16} mode. Key: A (Attention) indicates the use of attention weights; H (Hidden states) indicates the use of hidden states. Note that the Entropy method uses neither Attention nor Hidden states, and thus represents the lowest memory consumption required.}
    \label{tab:memory_consumption}
\end{table}

\paragraph{Benchmarks.} We use the MSCOCO~\cite{lin_microsoft_2014} dataset following previous works~\cite{jiang2025devils,park2025glsim}, and additionally the Pascal VOC~\cite{everingham_pascal_2010} dataset. For the MSCOCO dataset, following~\cite{park2025glsim}, we randomly sample 5,000 images from the \texttt{val2014} subset, and for the Pascal VOC dataset we use the full \texttt{val2012} subset, which contains 5,823 images. The MSCOCO and Pascal VOC dataset contain 80 and 20 object classes, respectively. These benchmarks contain, for each image, a list of objects known to be present in that image, and string matching against this list is used to discover object tokens, same as in CHAIR~\cite{rohrbach2018chair} hallucination evaluation. All models are prompted with ``\textit{Please help me describe the image in detail.}'' 

\paragraph{Baselines.} We compare our detection method against other baselines, including \textit{Logits}-based approaches: Negative Log-Likelihood (NLL)~\cite{zhou2024analyzing} and Entropy~\cite{malinin2021uncertainty}; \textit{Attention}-based approach: Summed Visual Attention Ratio (SVAR)~\cite{jiang2025devils}; and \textit{Representation}-based approaches: Internal Confidence (IC)~\cite{jiang_interpreting_2024} and Global-Local similarity (GLSim)~\cite{park2025glsim}. All baselines are evaluated on the same samples per dataset. For parameterized baselines, we use the parameters provided in the original papers for the models evaluated, if exists.

\subsection{Results}
\label{sec:results}

\begin{table}[t]
    \centering
    
    \resizebox{.92\linewidth}{!}{
    \begin{tabular}{l c c c c c c | c} 
        \toprule
        \textbf{Method} & \textbf{Greedy} & \textbf{Beam search} & \textbf{Top-k} & \textbf{Nucleus} \\
        \midrule
        % IC & XX.X & XX.X & XX.X & XX.X & XX.X & XX.X & XX.X \\
        NLL & 56.5 & 55.7 & 58.3 & 59.9 \\
        Entropy & 71.6 & 71.9 & 74.0 & 75.3 \\
        IC & 75.1 & 73.0 & 76.7 & 77.9 \\
        GLSim & 64.1 & 65.8 & 65.5 & 65.4 \\
        SVAR & 81.5 & 81.3 & 79.9 & 80.4 \\
        Ours & \textbf{84.2} & \textbf{84.0} & \textbf{83.5} & \textbf{84.0} \\
        \bottomrule
    \end{tabular}
    }
    \caption{Effect of decoding methods on object hallucination detection performance (AUROC) for LLaVA-1.5-7B on MSCOCO. All values are percentages, and best results are shown in \textbf{bold}.}
    \label{tab:decoding_strategies}
\end{table}

In \Cref{tab:induce_score}, we compare different realizations of \cref{eq:induce_score} in terms of detection performance against the attention-based PAS. From this result, it can be seen that our information-theoretic formulation (\cref{eq:induce_entropy_diff}) leads to efficacious object hallucination detection, albeit with different performance depending on the choice of $\Delta$. We note with interest that our simple PAS method not only proves far more efficient (requiring no extra forward passes) but also achieves superior detection performance compared to all $D_{\Delta}$ variants. 
We hypothesize this is because PAS directly measures the model's internal information dependency for the current, specific input. 
In contrast, the $D_{\Delta}$ approach (\cref{eq:induce_score}) is limited by the necessary but noisy estimation of the marginal probability distribution $\Pr(y_{k} \mid \mathbf{y}_{<k}, \mathbf{t})$ across a finite set of reference images $\mathcal{I}$. 
This suggests that direct attention flow is a cleaner and more robust signal of the model's ``alternative operating mode'' than distributional differences. 
Ultimately, the fact that both the probability-based $D_{\Delta}$ methods and the attention-based PAS yield effective detectors strongly supports our hypothesis \textbf{H1}: that prelim overdependence plays a significant role in object hallucination in LVLMs.

\Cref{fig:qualitative} illustrates a qualitative example of our PAS method. In particular, we compare the attention from suffixes of the prelim to the two object tokens. The first object token receives more attention from the prelim, which indicates higher chance of hallucination. Notably, for the hallucinatory token, the attention seems to concentrate at a few tokens, while for the non-hallucinatory token it is more dispersed, which is potentially related to observations by \cite{huang2024opera}.

% While different choices of $\Delta$ lead to varying degree of effectiveness, most of them are competitive across models and datasets.

\begin{figure}[t]
    \centering
    \begin{subfigure}[t]{0.48\linewidth}
        \centering
        \includegraphics[width=\linewidth]{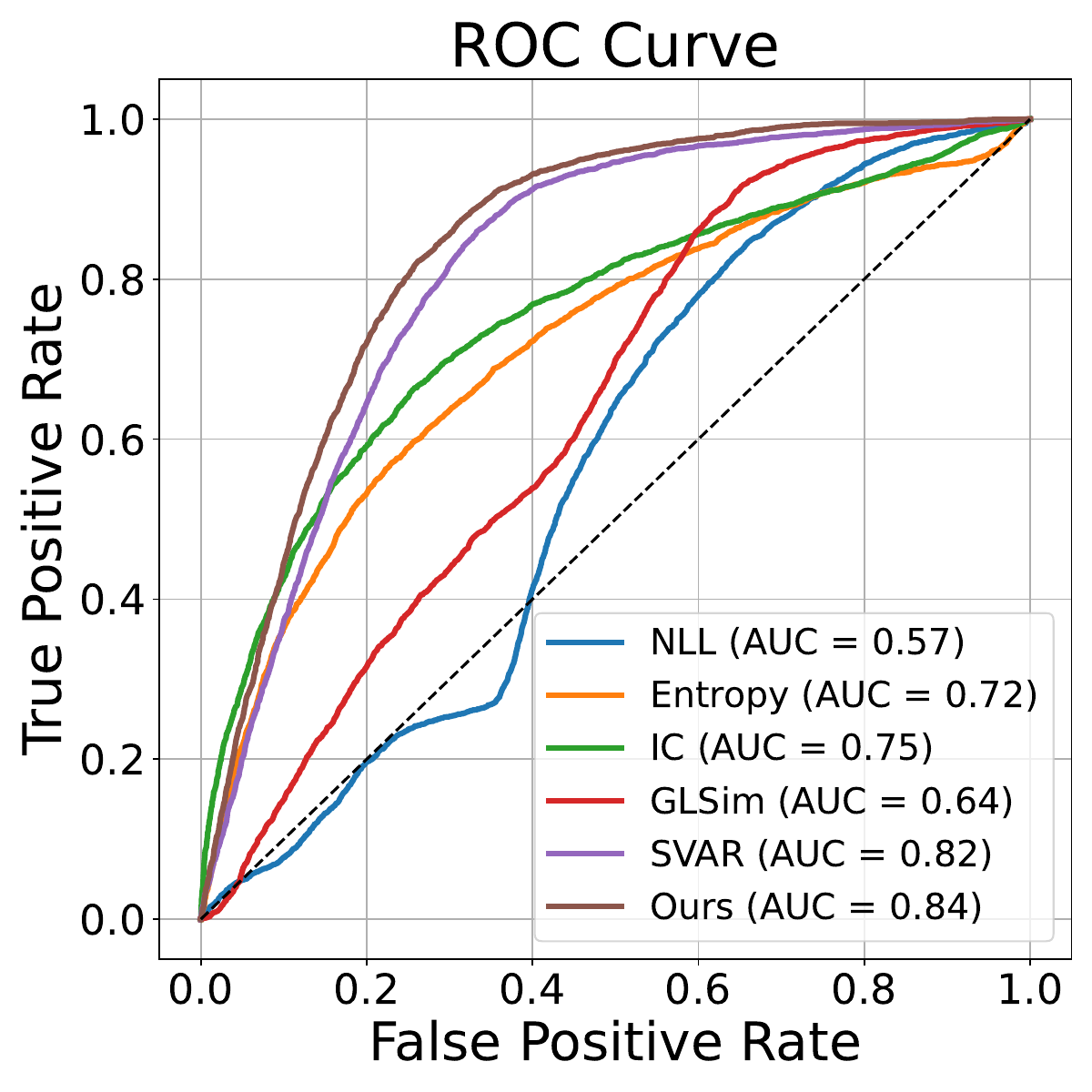}
        % \caption{PAS score (\cref{eq:prelim_attention_score})}
        % \label{fig:roc_curve}
    \end{subfigure}%
    ~ 
    \begin{subfigure}[t]{0.48\linewidth}
        \centering
        \includegraphics[width=\linewidth]{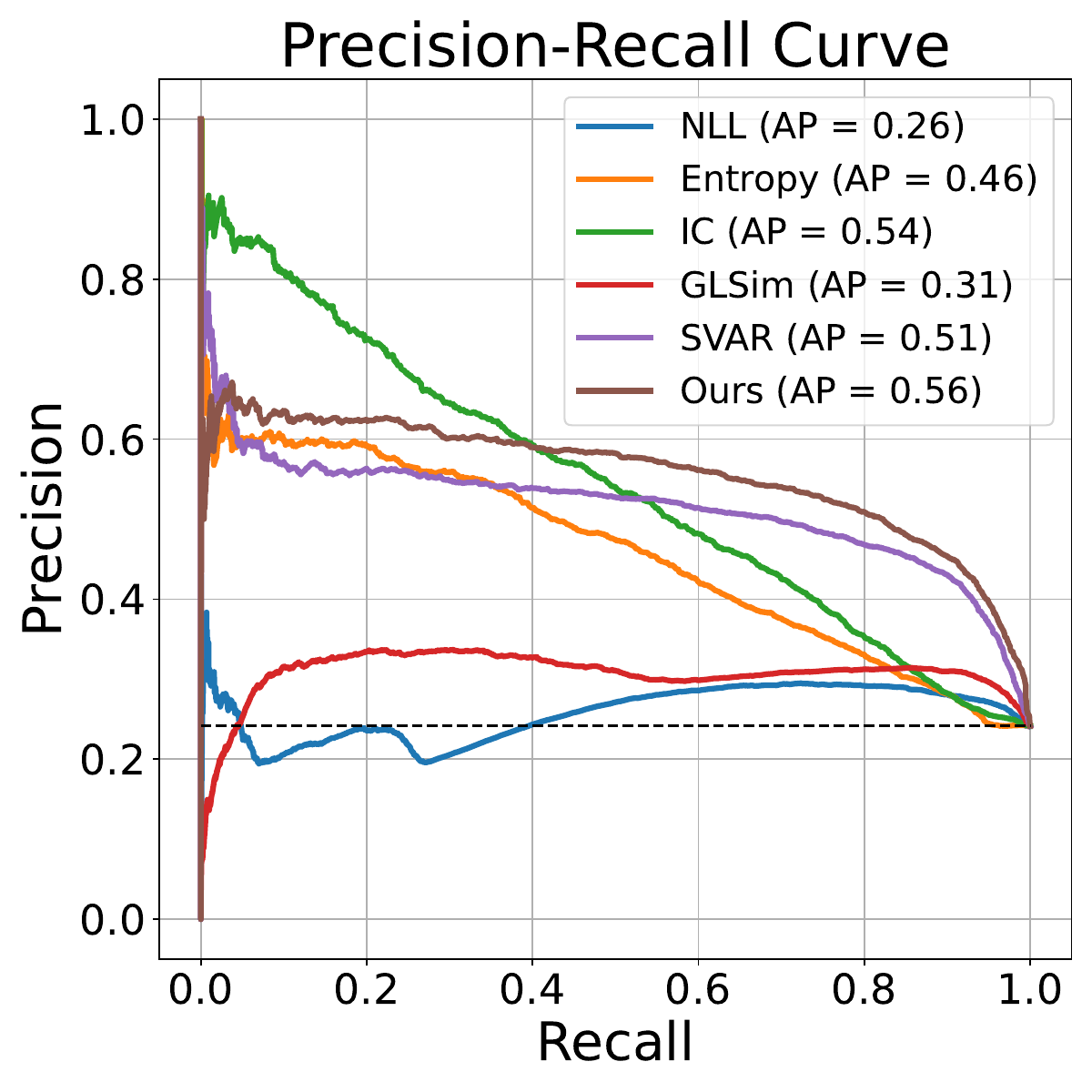}
        % \caption{MI-based score (\cref{eq:logit_induce_score})}
        % \label{fig:prc_curve}
    \end{subfigure}
    \caption{The ROC and PRC curves for object hallucination detection of our method and the baselines for LLaVA-1.5-7B on MSCOCO dataset. Dashed line indicates chance performance. }
    \label{fig:roc_prc_curves}
    % \label{fig:violin}
\end{figure}

\Cref{tab:main_results} describes the main comparative study against other baselines on object hallucination detection. From the table, our PAS consistently outperforms the baselines across most models and datasets, highlighting the usefulness of prelim tokens attention for detection. Interestingly, the most competitive baseline is SVAR~\cite{jiang2025devils}, which is also entirely attention-based, but they focus on image tokens instead. For a more detailed comparison, we plot the Receiver operating characteristic curve (ROC) and Precision-recall curve (PRC) of different methods for LLaVA-1.5-7B on MSCOCO in \Cref{fig:roc_prc_curves}. The curves suggest that our method remains effective irrespective of the choice of the threshold $\tau$. Additionally, \Cref{fig:violin} depicts the distribution of our score for real and hallucinated tokens, showing that their distributions are effectively separated, as desired.

Finally, \Cref{tab:memory_consumption} indicates that our method incurs minimal GPU memory  overhead, being on par with attention-based SVAR~\cite{jiang2025devils} and using 33\% less \emph{extra} memory per sample than hidden states-based GLSim~\cite{park2025glsim}, highlighting our method's efficiency.

\subsection{Ablation studies}
\label{sec:ablation_study}

\begin{figure}[t]
    \centering
    \includegraphics[width=0.95\linewidth]{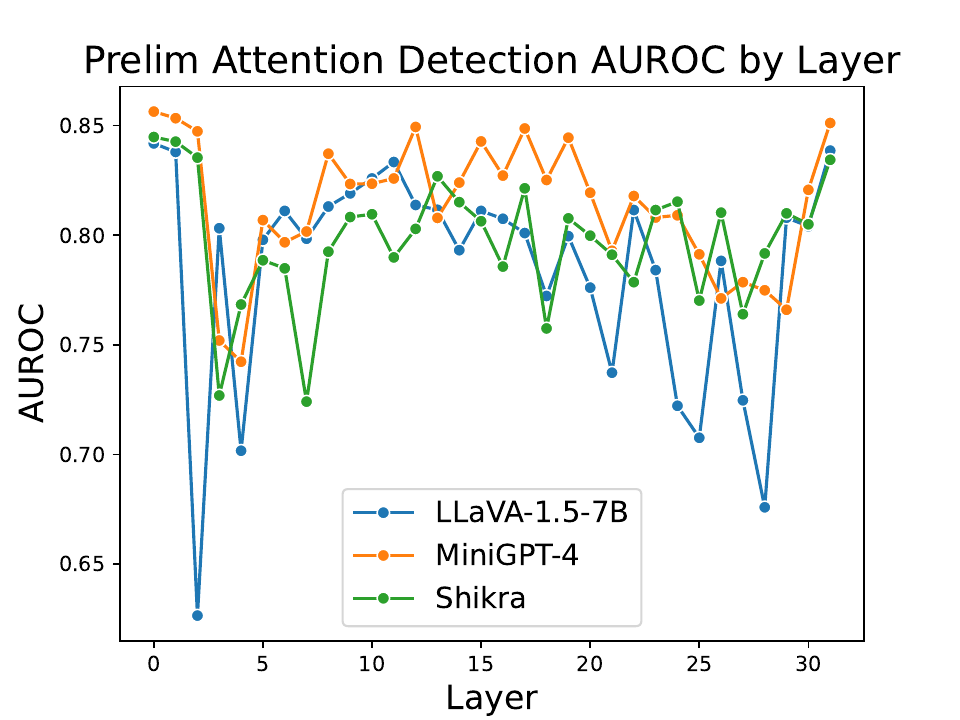}
    \caption{Comparison of different layers' performance in detecting object hallucination on the MSCOCO dataset. }
    \label{fig:ablate_layer}
\end{figure}

\paragraph{Layer $l$ for computing attention score.} We perform an ablation test on the layer $l$ (\cref{eq:prelim_attention_score}) chosen to compute PAS and visualize the results in \Cref{fig:ablate_layer}. Notably, the first layer tends to give the best results. Interestingly, this result aligns with existing research~\cite{artzy_attend_2024,gupta_how_2025} suggesting that earlier layers focus on gathering information and latter layers focus on processing that information. Based on this, we adopt the first layer (layer 0) to compute the attention scores for our method for the models evaluated in this paper.

\paragraph{Attention from token types other than prelim.} 

% Experimentally, we find that $s_\text{BOS},s_\text{ins},s_\text{img}$ have negative correlation with hallucination, while $s_\text{prel}$ has positive correlation with hallucination. Their detection performance is depicted in \Cref{tab:ablate_attention_types}. We display both global (i.e., averaged over all layers) and layer 0 variants for a better overall picture. Note that existing methods have leveraged some of them, for example, SVAR~\cite{jiang2025devils} sums image attention (to object token) over middle layers, which is similar to $s_\text{img}$. Nonetheless, \Cref{tab:ablate_attention_types} shows that the prelim attention score is most effective overall.

Experimentally, we find that $s_\text{BOS},s_\text{ins},s_\text{img}$ have negative correlation with hallucination, while $s_\text{prel}$ has positive correlation with hallucination. Their detection performance is depicted in \Cref{tab:ablate_attention_types}. We display both global (i.e., averaged over all layers) and layer 0 variants for a better overall picture. 

\begin{figure}[t]
    \centering % Center the figure
    
    \begin{subfigure}[b]{0.48\linewidth}
        \centering
        \includegraphics[width=\linewidth]{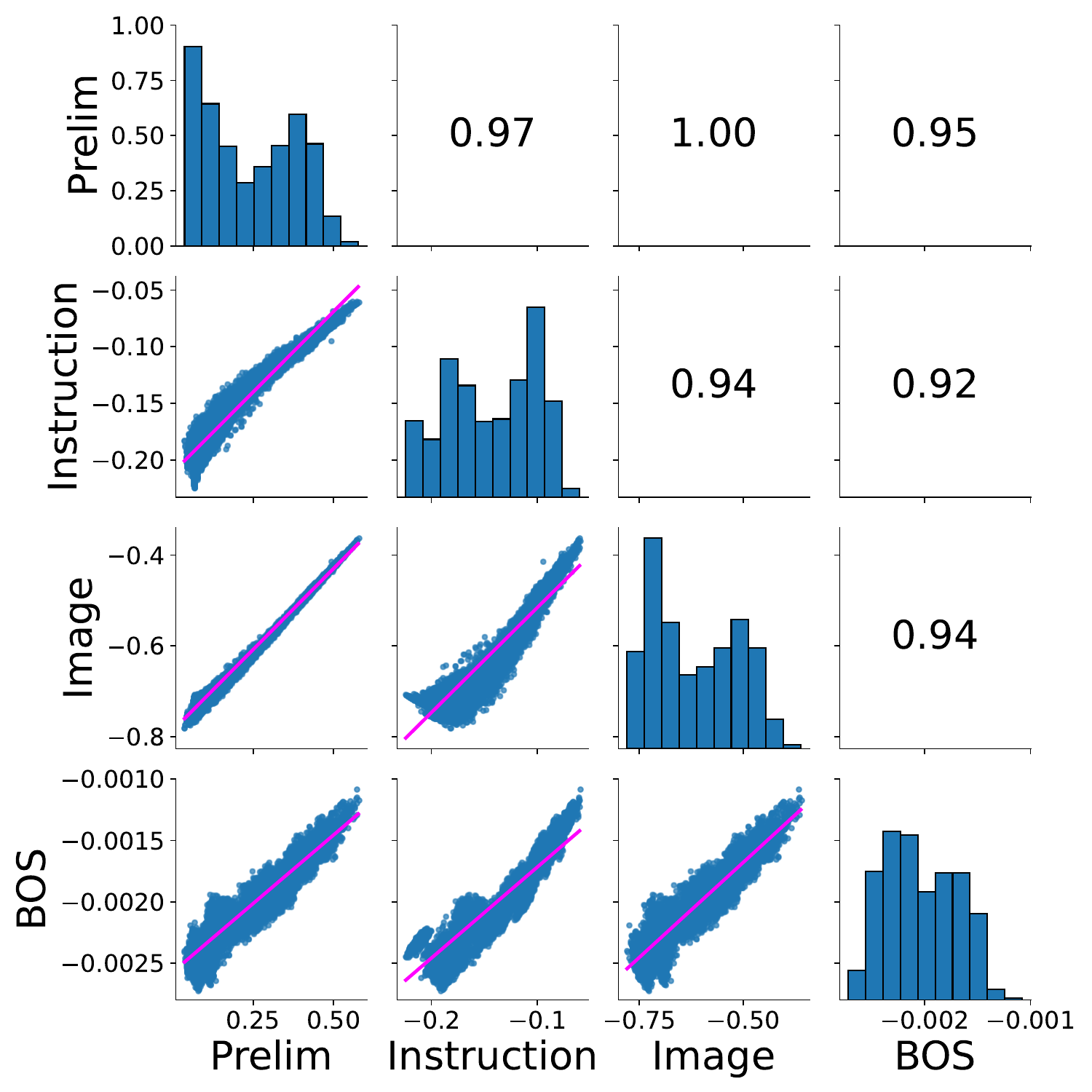}
        \caption{LLaVA-1.5-7B}
        % \label{fig:first}
    \end{subfigure}
    \hfill % Adds horizontal space between the figures
    \begin{subfigure}[b]{0.48\linewidth}
        \centering
        \includegraphics[width=\linewidth]{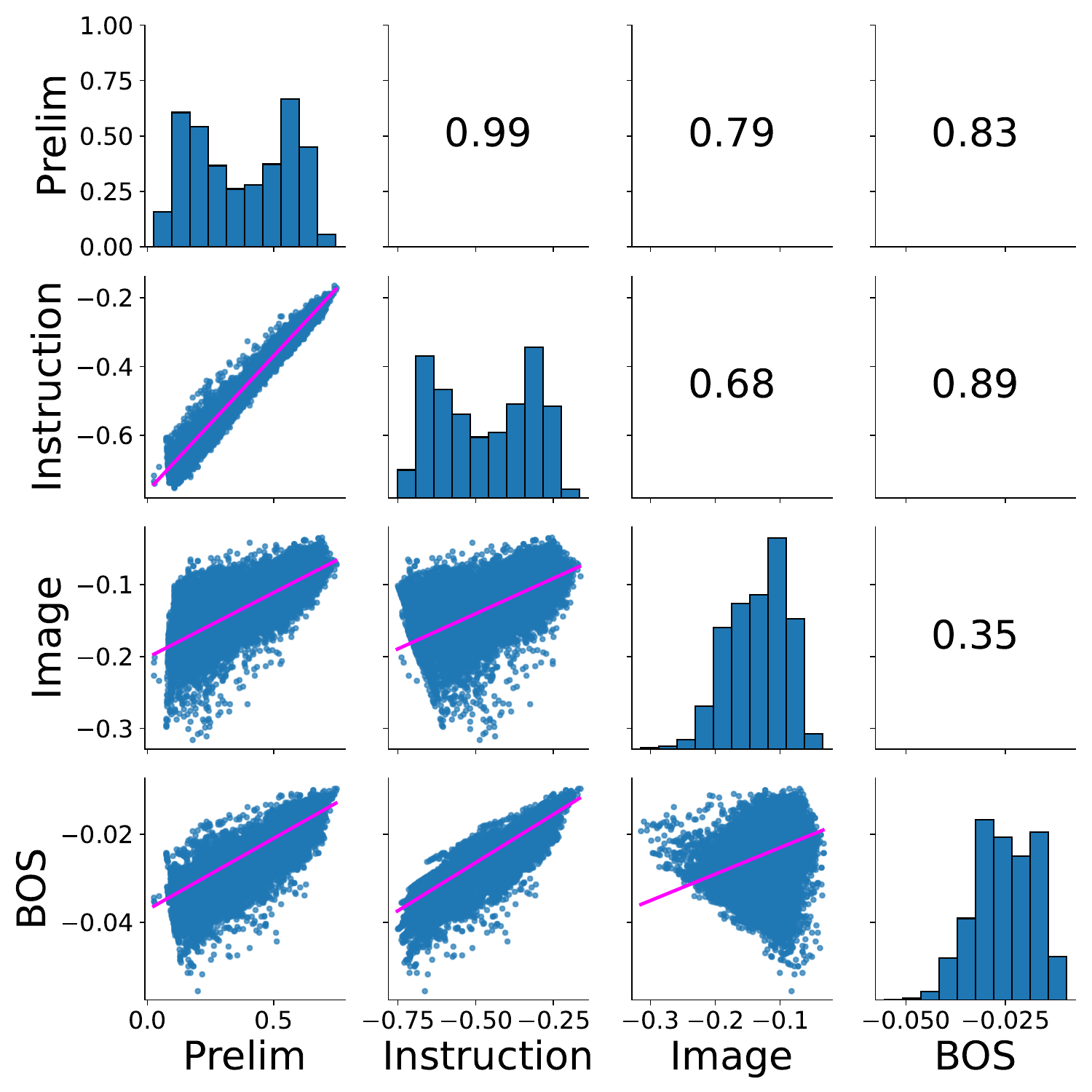}
        \caption{Shikra}
        % \label{fig:second}
    \end{subfigure}
    
    \caption{Correlation between attention scores (layer 0) from different token types to object tokens for LLaVA-1.5-7B and Shikra on MSCOCO. Upper-right half shows Pearson correlation coefficients. All scores other than prelim attention score are negated so that high score indicates high likelihood of hallucination. }
    \label{fig:attn_cor}
\end{figure}

A key question is whether PAS ($s_\text{prel}$) offers a genuine advantage over simply using image attention ($s_\text{img}$), which is the basis for methods like SVAR~\cite{jiang2025devils}. The softmax constraint in self-attention causes these scores to be correlated (as visualized in \Cref{fig:attn_cor}). However, our results demonstrate they are not redundant signals. \Cref{tab:ablate_attention_types} clearly shows that $s_\text{prel}$ from layer 0 (84.8\% avg) provides a more effective detection signal than $s_\text{img}$ from the same layer (82.1\% avg). This suggests that while low image attention is a \textit{symptom} of hallucination, high prelim attention is a more direct and potent \textit{signal} of the model's shift to an unreliable ``alternative operating mode''. 

Note that \Cref{tab:ablate_attention_types} also shows that some other attention scores (namely instruction tokens) exhibit significant detection capabilities, while not necessarily admitting a theoretically-inspired explanation similar to what we formalized for the prelim tokens. We theorize that this is because the softmax operation in the self-attention layers causes some degree of correlation between different attention scores. We visualize the pairwise correlation between the scores for two different models in \Cref{fig:attn_cor}. Indeed, for the models we considered, there is significant correlation between the prelim attention and instruction attention, which can explain why the instruction attention score performs well in \Cref{tab:ablate_attention_types}.

% A prompt typically consists of BOS, instruction, image, and prelim tokens, and the sum of their attention to an object token is constrained to be equal to one by the softmax function. This leads to significant correlation between different attention scores for a single model, as visualized in \Cref{fig:attn_cor}. We experiment with using different attention scores for hallucination detection and report the result in \Cref{tab:ablate_attention_types}. 
% Note that high correlation does not imply all attention scores are equally effective for detection, and we find the prelim attention score to be most effective and reliable overall.

\paragraph{Effect of decoding strategies.} Decoding schemes can significantly affect hallucination rates of LVLMs~\cite{huang2024opera}. Following existing works~\cite{park2025glsim,jiang2025devils}, we use greedy decoding as the default evaluation setting due to its computational efficiency and simplicity. Here, we additionally investigate the effect of different decoding schemes on detection performance. We run the LVLMs with three additional distinct decoding strategies and perform the same quantitative experiments on object hallucination detection. We use $N_\text{beams}=5$, $k=10$, and $p=0.9$ for beam search, top-k decoding, and nucleus decoding, respectively, and report the results in \Cref{tab:decoding_strategies}. Results suggest that our method consistently provides a significant gain in detection performance irrespective of the decoding scheme used, ensuring its practicality in real-world scenarios where decoding schemes can change depending on the task at hand~\cite{shi_thorough_2024}.

\begin{table}[t]
    \centering
    \resizebox{.92\linewidth}{!}{
    \begin{tabular}{l c c c | c} 
        \toprule
        \textbf{Method} & \textbf{LLaVA-1.5-7B} & \textbf{MiniGPT-4} & \textbf{Shikra} & \textbf{Average} \\
        \midrule
        \multicolumn{4}{l}{\textit{Global attention scores}} \\
        \midrule
        Prelim       & 83.7 & 86.2 & 84.4 & 84.8 \\
        Instruction  & 79.7 & 86.3 & 84.6 & 83.5 \\        
        Image        & 75.0 & 88.7 & 72.4 & 78.7 \\
        BOS          & 74.0 & 74.9 & 55.5 & 68.1 \\
        \midrule
        \multicolumn{4}{l}{\textit{Layer 0 attention scores}} \\
        \midrule
        Prelim      & 84.2 & 85.6 & 84.5 & 84.8 \\
        Instruction & 83.9 & 85.7 & 84.5 & 84.7 \\
        Image       & 84.1 & 85.3 & 77.0 & 82.1 \\
        BOS         & 83.1 & 85.4 & 82.1 & 83.5 \\
        \bottomrule
    \end{tabular}
    }
    \caption{Hallucination detection results (AUROC) when using different types of attention scores on MSCOCO. \textit{Global} score means averaging the attention weights over all heads in a layer and over all layers in a model. }
    \label{tab:ablate_attention_types}
\end{table}

\section{Conclusion}
\label{sec:conclusion}

We presented PAS, a training-free detector for object hallucinations in LVLMs that exploits a previously overlooked cue: \emph{attention flow from preliminary, low-information output tokens to object tokens}. We support our method with an information-theoretic interpretation that links high prelim attention with an alternative unreliable operating mode of the LVLM. Experiments across models and datasets show that PAS achieves state-of-the-art object hallucination detection performance while being computationally efficient, requiring no extra passes. Our findings complement the literature on LVLM object hallucination, suggesting the existence of an undesired operating mode of the model and its detection, thus paving the way for future development of more dependable and trustworthy LVLMs.

% By quantifying this \emph{prelim attention} (Eq.~(1)) and optionally coupling it with a lightweight probability-contrast proxy (Eq.~(2)), PAL offers an interpretable ``where the model looked before speaking'' lens that complements probability-, visual-attention-, and hidden-state methods. Across LLaVA-1.5, InstructBLIP, and MiniGPT-4, PAL achieves state-of-the-art detection on POPE, CHAIR, and MMHal-style evaluations in a single forward pass, with $<\!5\%$ overhead and no retraining or auxiliary models. Analysis shows (i) early and last layers carry the strongest signal, (ii) attention-type variants are correlated, and (iii) hallucinations exhibit a consistent \emph{attention leakage} pattern toward BOS/system/format tokens when visual grounding is weak. PAL turns readily available inference-time signals into actionable diagnostics and controls. We hope its simplicity, interpretability, and strong empirical performance make it a practical default for hallucination detection and a probe for guiding future LVLM architectures and training schemes toward persistent visual grounding.

\paragraph{Limitations and future work.}
Our observations for the PAS score are based on the original self-attention proposed in~\cite{vaswani_attention_2017} and is widely used by many LVLMs/LLMs. Besides full self-attention, other variants~\cite{abdin_phi-3_2024,beltagy_longformer_2020} have been proposed to reduce the quadratic complexity of self-attention. For those variants, since the output token might no longer attend to all token types, our hypothesis/observations might not hold. Moreover, in this work, we focus on object existence hallucination; other types of hallucination such as relational/attribute hallucination are left for future work. Finally, our experiments focused on open-ended image description. Future work should investigate whether PAS remains an effective signal in other tasks, such as Visual Question Answering (VQA). It is plausible that in VQA, the varying questions introduce different prelim token dynamics, providing an interesting avenue for investigations.

\section*{Acknowledgment}
This manuscript has been assigned LA-UR-25-30786. This research was funded by the Los Alamos National Laboratory (LANL) Laboratory Directed Research and Development (LDRD) program under grants 20250850ECR and 20240868PRD3 and supported by LANL’s Institutional Computing Program, and by the U.S. Department of Energy National Nuclear Security Administration under Contract No. 89233218CNA000001. 
{
    \small
    \bibliographystyle{ieeenat_fullname}
    \bibliography{main}

@String(CVPR= {IEEE Conf. Comput. Vis. Pattern Recog.})

@String(ECCV= {Eur. Conf. Comput. Vis.})

@String(ICLR = {Int. Conf. Learn. Represent.})

@String(AAAI = {AAAI})

@String(CVPR  = {CVPR})

@String(ECCV  = {ECCV})

@String(ICLR  = {ICLR})

@inproceedings{park2025glsim,
    title = {{GLSim}: {Detecting} object hallucinations in {LVLMs} via global-local similarity},
    url = {https://openreview.net/forum?id=riFkEuOIcp},
    booktitle = {{ICML} 2025 workshop on reliable and responsible foundation models},
    author = {Park, Seongheon and Li, Yixuan},
    year = {2025},
}

@inproceedings{jiang_interpreting_2024,
    title = {Interpreting and {Editing} {Vision}-{Language} {Representations} to {Mitigate} {Hallucinations}},
    url = {https://openreview.net/forum?id=94kQgWXojH},
    language = {en},
    urldate = {2025-10-27},
    author = {Jiang, Nicholas and Kachinthaya, Anish and Petryk, Suzanne and Gandelsman, Yossi},
    month = oct,
    year = {2024},
}

@inproceedings{jiang2025devils,
    title = {Devils in middle layers of large vision-language models: {Interpreting}, detecting and mitigating object hallucinations via attention lens},
    booktitle = {Proceedings of the computer vision and pattern recognition conference},
    author = {Jiang, Zhangqi and Chen, Junkai and Zhu, Beier and Luo, Tingjin and Shen, Yankun and Yang, Xu},
    year = {2025},
    pages = {25004--25014},
}

@inproceedings{malinin2021uncertainty,
    title = {Uncertainty estimation in autoregressive structured prediction},
    url = {https://openreview.net/forum?id=jN5y-zb5Q7m},
    booktitle = {International conference on learning representations},
    author = {Malinin, Andrey and Gales, Mark},
    year = {2021},
}

@misc{gupta_how_2025,
    title = {How {Do} {LLMs} {Use} {Their} {Depth}?},
    url = {http://arxiv.org/abs/2510.18871},
    doi = {10.48550/arXiv.2510.18871},
    urldate = {2025-10-27},
    publisher = {arXiv},
    author = {Gupta, Akshat and Yeung, Jay and Anumanchipalli, Gopala and Ivanova, Anna},
    month = oct,
    year = {2025},
    note = {arXiv:2510.18871},
}

@inproceedings{liu_visual_2023,
    title = {Visual {Instruction} {Tuning}},
    volume = {36},
    url = {https://proceedings.neurips.cc/paper_files/paper/2023/file/6dcf277ea32ce3288914faf369fe6de0-Paper-Conference.pdf},
    booktitle = {Advances in {Neural} {Information} {Processing} {Systems}},
    publisher = {Curran Associates, Inc.},
    author = {Liu, Haotian and Li, Chunyuan and Wu, Qingyang and Lee, Yong Jae},
    editor = {Oh, A. and Naumann, T. and Globerson, A. and Saenko, K. and Hardt, M. and Levine, S.},
    year = {2023},
    pages = {34892--34916},
}

@inproceedings{zhu2024minigpt,
    title = {{MiniGPT}-4: {Enhancing} vision-language understanding with advanced large language models},
    url = {https://openreview.net/forum?id=1tZbq88f27},
    booktitle = {The twelfth international conference on learning representations},
    author = {Zhu, Deyao and Chen, Jun and Shen, Xiaoqian and Li, Xiang and Elhoseiny, Mohamed},
    year = {2024},
}

@article{chen_shikra_2023,
    title = {Shikra: {Unleashing} multimodal llm's referential dialogue magic},
    journal = {arXiv preprint arXiv:2306.15195},
    author = {Chen, Keqin and Zhang, Zhao and Zeng, Weili and Zhang, Richong and Zhu, Feng and Zhao, Rui},
    year = {2023},
}

@inproceedings{lin_microsoft_2014,
    series = {Lecture {Notes} in {Computer} {Science}},
    title = {Microsoft {COCO}: {Common} {Objects} in {Context}},
    volume = {8693},
    shorttitle = {Microsoft {COCO}},
    url = {https://doi.org/10.1007/978-3-319-10602-1\_48},
    doi = {10.1007/978-3-319-10602-1_48},
    urldate = {2022-05-05},
    booktitle = {Computer {Vision} - {ECCV} 2014 - 13th {European} {Conference}, {Zurich}, {Switzerland}, {September} 6-12, 2014, {Proceedings}, {Part} {V}},
    publisher = {Springer},
    author = {Lin, Tsung-Yi and Maire, Michael and Belongie, Serge J. and Hays, James and Perona, Pietro and Ramanan, Deva and Dollár, Piotr and Zitnick, C. Lawrence},
    editor = {Fleet, David J. and Pajdla, Tomás and Schiele, Bernt and Tuytelaars, Tinne},
    year = {2014},
    pages = {740--755},
}

@article{everingham_pascal_2010,
    title = {The {Pascal} {Visual} {Object} {Classes} ({VOC}) {Challenge}},
    volume = {88},
    issn = {1573-1405},
    url = {https://doi.org/10.1007/s11263-009-0275-4},
    doi = {10.1007/s11263-009-0275-4},
    language = {en},
    number = {2},
    urldate = {2025-10-27},
    journal = {International Journal of Computer Vision},
    author = {Everingham, Mark and Van Gool, Luc and Williams, Christopher K. I. and Winn, John and Zisserman, Andrew},
    month = jun,
    year = {2010},
    pages = {303--338},
}

@inproceedings{zhou2024analyzing,
  title     = {Analyzing and Mitigating Object Hallucination in Large Vision-Language Models},
  author    = {Zhou, Yiyang and Cui, Chenhang and Yoon, Jaehong and Zhang, Linjun and Deng, Zhun and Finn, Chelsea and Bansal, Mohit and Yao, Huaxiu},
  booktitle = {The Twelfth International Conference on Learning Representations (ICLR)},
  year      = {2024},
  url       = {https://openreview.net/forum?id=JXGJQluKxS}
}

@inproceedings{huang2024opera,
  title     = {OPERA: Over-trust Penalty with Rationale-Guided Decoding Mitigates Hallucination in Vision-Language Models},
  author    = {Huang, Jieru and Yang, Jinyue and Ma, Guangzhi and Chen, Danfeng and Kuen, Jason and Anandkumar, Anima and Huang, Gao},
  booktitle = {Proceedings of the IEEE/CVF Conference on Computer Vision and Pattern Recognition (CVPR)},
  pages     = {13418--13427},
  year      = {2024},
  url       = {https://openaccess.thecvf.com/content/CVPR2024/html/Huang_OPERA_Over-trust_Penalty_with_Rationale-Guided_Decoding_Mitigates_Hallucination_in_Vision-Language_CVPR_2024_paper.html}
}

@inproceedings{leng2024vcd,
  title     = {Mitigating Object Hallucinations in Large Vision-Language Models through Visual Contrastive Decoding},
  author    = {Leng, Sicong and Wang, Jingjing and Dong, Jinming and Gao, Yang and Wang, Weizhi and Lim, Ee-Peng and Bing, Lidong},
  booktitle = {Proceedings of the IEEE/CVF Conference on Computer Vision and Pattern Recognition (CVPR)},
  year      = {2024},
  url       = {https://openaccess.thecvf.com/content/CVPR2024/html/Leng_Mitigating_Object_Hallucinations_in_Large_Vision-Language_Models_through_Visual_Contrastive_CVPR_2024_paper.html}
}

@inproceedings{li2023pope,
  title     = {Evaluating Object Hallucination in Large Vision-Language Models},
  author    = {Li, Yifan and Du, Yifan and Zhou, Kun and Wang, Jinpeng and Zhao, Wayne Xin and Wen, Ji-Rong},
  booktitle = {Proceedings of the 2023 Conference on Empirical Methods in Natural Language Processing (EMNLP)},
  pages     = {292--305},
  address   = {Singapore},
  year      = {2023},
  publisher = {Association for Computational Linguistics},
  url       = {https://aclanthology.org/2023.emnlp-main.20/}
}

@inproceedings{rohrbach2018chair,
  title     = {Object Hallucination in Image Captioning},
  author    = {Rohrbach, Anna and Hendricks, Lisa Anne and Burns, Kaylee and Darrell, Trevor and Saenko, Kate},
  booktitle = {Proceedings of the 2018 Conference on Empirical Methods in Natural Language Processing (EMNLP)},
  pages     = {4035--4045},
  year      = {2018},
  publisher = {Association for Computational Linguistics},
  url       = {https://aclanthology.org/D18-1437/}
}

@inproceedings{an2025mitigating,
  title={Mitigating object hallucinations in large vision-language models with assembly of global and local attention},
  author={An, Wenbin and Tian, Feng and Leng, Sicong and Nie, Jiahao and Lin, Haonan and Wang, QianYing and Chen, Ping and Zhang, Xiaoqin and Lu, Shijian},
  booktitle={Proceedings of the Computer Vision and Pattern Recognition Conference},
  pages={29915--29926},
  year={2025}
}

@misc{touvron_llama_2023,
    title = {Llama 2: {Open} {Foundation} and {Fine}-{Tuned} {Chat} {Models}},
    shorttitle = {Llama 2},
    url = {http://arxiv.org/abs/2307.09288},
    doi = {10.48550/arXiv.2307.09288},
    urldate = {2024-01-18},
    publisher = {arXiv},
    author = {Touvron, Hugo and Martin, Louis and Stone, Kevin and Albert, Peter and Almahairi, Amjad and Babaei, Yasmine and Bashlykov, Nikolay and Batra, Soumya and Bhargava, Prajjwal and Bhosale, Shruti and Bikel, Dan and Blecher, Lukas and Ferrer, Cristian Canton and Chen, Moya and Cucurull, Guillem and Esiobu, David and Fernandes, Jude and Fu, Jeremy and Fu, Wenyin and Fuller, Brian and Gao, Cynthia and Goswami, Vedanuj and Goyal, Naman and Hartshorn, Anthony and Hosseini, Saghar and Hou, Rui and Inan, Hakan and Kardas, Marcin and Kerkez, Viktor and Khabsa, Madian and Kloumann, Isabel and Korenev, Artem and Koura, Punit Singh and Lachaux, Marie-Anne and Lavril, Thibaut and Lee, Jenya and Liskovich, Diana and Lu, Yinghai and Mao, Yuning and Martinet, Xavier and Mihaylov, Todor and Mishra, Pushkar and Molybog, Igor and Nie, Yixin and Poulton, Andrew and Reizenstein, Jeremy and Rungta, Rashi and Saladi, Kalyan and Schelten, Alan and Silva, Ruan and Smith, Eric Michael and Subramanian, Ranjan and Tan, Xiaoqing Ellen and Tang, Binh and Taylor, Ross and Williams, Adina and Kuan, Jian Xiang and Xu, Puxin and Yan, Zheng and Zarov, Iliyan and Zhang, Yuchen and Fan, Angela and Kambadur, Melanie and Narang, Sharan and Rodriguez, Aurelien and Stojnic, Robert and Edunov, Sergey and Scialom, Thomas},
    month = jul,
    year = {2023},
    note = {arXiv:2307.09288 [cs]},
}

@inproceedings{liu_paying_2024,
    address = {Berlin, Heidelberg},
    title = {Paying {More} {Attention} to {Image}: {A} {Training}-{Free} {Method} for {Alleviating} {Hallucination} in {LVLMs}},
    isbn = {978-3-031-73009-2},
    shorttitle = {Paying {More} {Attention} to {Image}},
    url = {https://doi.org/10.1007/978-3-031-73010-8_8},
    doi = {10.1007/978-3-031-73010-8_8},
    urldate = {2025-11-01},
    booktitle = {Computer {Vision} – {ECCV} 2024: 18th {European} {Conference}, {Milan}, {Italy}, {September} 29–{October} 4, 2024, {Proceedings}, {Part} {LXXXIII}},
    publisher = {Springer-Verlag},
    author = {Liu, Shi and Zheng, Kecheng and Chen, Wei},
    month = nov,
    year = {2024},
    pages = {125--140},
}

@misc{bai_qwen_2023,
    title = {Qwen {Technical} {Report}},
    url = {http://arxiv.org/abs/2309.16609},
    doi = {10.48550/arXiv.2309.16609},
    urldate = {2025-11-02},
    publisher = {arXiv},
    author = {Bai, Jinze and Bai, Shuai and Chu, Yunfei and Cui, Zeyu and Dang, Kai and Deng, Xiaodong and Fan, Yang and Ge, Wenbin and Han, Yu and Huang, Fei and Hui, Binyuan and Ji, Luo and Li, Mei and Lin, Junyang and Lin, Runji and Liu, Dayiheng and Liu, Gao and Lu, Chengqiang and Lu, Keming and Ma, Jianxin and Men, Rui and Ren, Xingzhang and Ren, Xuancheng and Tan, Chuanqi and Tan, Sinan and Tu, Jianhong and Wang, Peng and Wang, Shijie and Wang, Wei and Wu, Shengguang and Xu, Benfeng and Xu, Jin and Yang, An and Yang, Hao and Yang, Jian and Yang, Shusheng and Yao, Yang and Yu, Bowen and Yuan, Hongyi and Yuan, Zheng and Zhang, Jianwei and Zhang, Xingxuan and Zhang, Yichang and Zhang, Zhenru and Zhou, Chang and Zhou, Jingren and Zhou, Xiaohuan and Zhu, Tianhang},
    month = sep,
    year = {2023},
    note = {arXiv:2309.16609 [cs]},
}

@article{gunjal_detecting_2024,
    title = {Detecting and {Preventing} {Hallucinations} in {Large} {Vision} {Language} {Models}},
    volume = {38},
    issn = {2374-3468, 2159-5399},
    url = {https://ojs.aaai.org/index.php/AAAI/article/view/29771},
    doi = {10.1609/aaai.v38i16.29771},
    number = {16},
    urldate = {2025-11-04},
    journal = {Proceedings of the AAAI Conference on Artificial Intelligence},
    author = {Gunjal, Anisha and Yin, Jihan and Bas, Erhan},
    month = mar,
    year = {2024},
    pages = {18135--18143},
}

@article{xiao_detecting_2025,
    title = {Detecting and {Mitigating} {Hallucination} in {Large} {Vision} {Language} {Models} via {Fine}-{Grained} {AI} {Feedback}},
    volume = {39},
    copyright = {Copyright (c) 2025 Association for the Advancement of Artificial Intelligence},
    issn = {2374-3468},
    url = {https://ojs.aaai.org/index.php/AAAI/article/view/34744},
    doi = {10.1609/aaai.v39i24.34744},
    language = {en},
    number = {24},
    urldate = {2025-11-04},
    journal = {Proceedings of the AAAI Conference on Artificial Intelligence},
    author = {Xiao, Wenyi and Huang, Ziwei and Gan, Leilei and He, Wanggui and Li, Haoyuan and Yu, Zhelun and Shu, Fangxun and Jiang, Hao and Zhu, Linchao},
    month = apr,
    year = {2025},
    pages = {25543--25551},
}

@inproceedings{li_reference-free_2024,
    address = {Miami, Florida, USA},
    title = {Reference-free {Hallucination} {Detection} for {Large} {Vision}-{Language} {Models}},
    url = {https://aclanthology.org/2024.findings-emnlp.262/},
    doi = {10.18653/v1/2024.findings-emnlp.262},
    urldate = {2025-11-04},
    booktitle = {Findings of the {Association} for {Computational} {Linguistics}: {EMNLP} 2024},
    publisher = {Association for Computational Linguistics},
    author = {Li, Qing and Geng, Jiahui and Lyu, Chenyang and Zhu, Derui and Panov, Maxim and Karray, Fakhri},
    editor = {Al-Onaizan, Yaser and Bansal, Mohit and Chen, Yun-Nung},
    month = nov,
    year = {2024},
    pages = {4542--4551},
}

@inproceedings{chen_unified_2024,
    address = {Bangkok, Thailand},
    title = {Unified {Hallucination} {Detection} for {Multimodal} {Large} {Language} {Models}},
    url = {https://aclanthology.org/2024.acl-long.178/},
    doi = {10.18653/v1/2024.acl-long.178},
    urldate = {2025-11-04},
    booktitle = {Proceedings of the 62nd {Annual} {Meeting} of the {Association} for {Computational} {Linguistics} ({Volume} 1: {Long} {Papers})},
    publisher = {Association for Computational Linguistics},
    author = {Chen, Xiang and Wang, Chenxi and Xue, Yida and Zhang, Ningyu and Yang, Xiaoyan and Li, Qiang and Shen, Yue and Liang, Lei and Gu, Jinjie and Chen, Huajun},
    editor = {Ku, Lun-Wei and Martins, Andre and Srikumar, Vivek},
    month = aug,
    year = {2024},
    pages = {3235--3252},
}

@article{yin_woodpecker_2024,
    title = {Woodpecker: hallucination correction for multimodal large language models},
    volume = {67},
    issn = {1869-1919},
    shorttitle = {Woodpecker},
    url = {https://doi.org/10.1007/s11432-024-4251-x},
    doi = {10.1007/s11432-024-4251-x},
    language = {en},
    number = {12},
    urldate = {2025-11-04},
    journal = {Science China Information Sciences},
    author = {Yin, Shukang and Fu, Chaoyou and Zhao, Sirui and Xu, Tong and Wang, Hao and Sui, Dianbo and Shen, Yunhang and Li, Ke and Sun, Xing and Chen, Enhong},
    month = dec,
    year = {2024},
    pages = {220105},
}

@inproceedings{jing_faithscore_2024,
    address = {Miami, Florida, USA},
    title = {{FaithScore}: {Fine}-grained {Evaluations} of {Hallucinations} in {Large} {Vision}-{Language} {Models}},
    shorttitle = {{FaithScore}},
    url = {https://aclanthology.org/2024.findings-emnlp.290/},
    doi = {10.18653/v1/2024.findings-emnlp.290},
    urldate = {2025-11-04},
    booktitle = {Findings of the {Association} for {Computational} {Linguistics}: {EMNLP} 2024},
    publisher = {Association for Computational Linguistics},
    author = {Jing, Liqiang and Li, Ruosen and Chen, Yunmo and Du, Xinya},
    editor = {Al-Onaizan, Yaser and Bansal, Mohit and Chen, Yun-Nung},
    month = nov,
    year = {2024},
    pages = {5042--5063},
}

@article{noauthor_interpreting_2020,
    title = {interpreting {GPT}: the logit lens — {LessWrong}},
    shorttitle = {interpreting {GPT}},
    url = {https://www.lesswrong.com/posts/AcKRB8wDpdaN6v6ru/interpreting-gpt-the-logit-lens},
    urldate = {2025-11-06},
    month = aug,
    year = {2020},
}

@inproceedings{gong_damro_2024,
    address = {Miami, Florida, USA},
    title = {{DAMRO}: {Dive} into the {Attention} {Mechanism} of {LVLM} to {Reduce} {Object} {Hallucination}},
    shorttitle = {{DAMRO}},
    url = {https://aclanthology.org/2024.emnlp-main.439/},
    doi = {10.18653/v1/2024.emnlp-main.439},
    urldate = {2025-11-08},
    booktitle = {Proceedings of the 2024 {Conference} on {Empirical} {Methods} in {Natural} {Language} {Processing}},
    publisher = {Association for Computational Linguistics},
    author = {Gong, Xuan and Ming, Tianshi and Wang, Xinpeng and Wei, Zhihua},
    editor = {Al-Onaizan, Yaser and Bansal, Mohit and Chen, Yun-Nung},
    month = nov,
    year = {2024},
    pages = {7696--7712},
}

@inproceedings{barbero_why_2025,
    title = {Why do {LLMs} attend to the first token?},
    url = {https://openreview.net/forum?id=tu4dFUsW5z#discussion},
    language = {en},
    urldate = {2025-11-08},
    author = {Barbero, Federico and Arroyo, Alvaro and Gu, Xiangming and Perivolaropoulos, Christos and Veličković, Petar and Pascanu, Razvan and Bronstein, Michael M.},
    month = aug,
    year = {2025},
}

@inproceedings{vaswani_attention_2017,
    title = {Attention is {All} you {Need}},
    url = {https://proceedings.neurips.cc/paper/2017/hash/3f5ee243547dee91fbd053c1c4a845aa-Abstract.html},
    urldate = {2022-05-04},
    booktitle = {Advances in {Neural} {Information} {Processing} {Systems} 30: {Annual} {Conference} on {Neural} {Information} {Processing} {Systems} 2017, {December} 4-9, 2017, {Long} {Beach}, {CA}, {USA}},
    author = {Vaswani, Ashish and Shazeer, Noam and Parmar, Niki and Uszkoreit, Jakob and Jones, Llion and Gomez, Aidan N. and Kaiser, Lukasz and Polosukhin, Illia},
    editor = {Guyon, Isabelle and Luxburg, Ulrike von and Bengio, Samy and Wallach, Hanna M. and Fergus, Rob and Vishwanathan, S. V. N. and Garnett, Roman},
    year = {2017},
    pages = {5998--6008},
}

@misc{abdin_phi-3_2024,
    title = {Phi-3 {Technical} {Report}: {A} {Highly} {Capable} {Language} {Model} {Locally} on {Your} {Phone}},
    shorttitle = {Phi-3 {Technical} {Report}},
    url = {http://arxiv.org/abs/2404.14219},
    doi = {10.48550/arXiv.2404.14219},
    urldate = {2025-11-09},
    publisher = {arXiv},
    author = {Abdin, Marah and Aneja, Jyoti and Awadalla, Hany and Awadallah, Ahmed and Awan, Ammar Ahmad and Bach, Nguyen and Bahree, Amit and Bakhtiari, Arash and Bao, Jianmin and Behl, Harkirat and Benhaim, Alon and Bilenko, Misha and Bjorck, Johan and Bubeck, Sébastien and Cai, Martin and Cai, Qin and Chaudhary, Vishrav and Chen, Dong and Chen, Dongdong and Chen, Weizhu and Chen, Yen-Chun and Chen, Yi-Ling and Cheng, Hao and Chopra, Parul and Dai, Xiyang and Dixon, Matthew and Eldan, Ronen and Fragoso, Victor and Gao, Jianfeng and Gao, Mei and Gao, Min and Garg, Amit and Giorno, Allie Del and Goswami, Abhishek and Gunasekar, Suriya and Haider, Emman and Hao, Junheng and Hewett, Russell J. and Hu, Wenxiang and Huynh, Jamie and Iter, Dan and Jacobs, Sam Ade and Javaheripi, Mojan and Jin, Xin and Karampatziakis, Nikos and Kauffmann, Piero and Khademi, Mahoud and Kim, Dongwoo and Kim, Young Jin and Kurilenko, Lev and Lee, James R. and Lee, Yin Tat and Li, Yuanzhi and Li, Yunsheng and Liang, Chen and Liden, Lars and Lin, Xihui and Lin, Zeqi and Liu, Ce and Liu, Liyuan and Liu, Mengchen and Liu, Weishung and Liu, Xiaodong and Luo, Chong and Madan, Piyush and Mahmoudzadeh, Ali and Majercak, David and Mazzola, Matt and Mendes, Caio César Teodoro and Mitra, Arindam and Modi, Hardik and Nguyen, Anh and Norick, Brandon and Patra, Barun and Perez-Becker, Daniel and Portet, Thomas and Pryzant, Reid and Qin, Heyang and Radmilac, Marko and Ren, Liliang and Rosa, Gustavo de and Rosset, Corby and Roy, Sambudha and Ruwase, Olatunji and Saarikivi, Olli and Saied, Amin and Salim, Adil and Santacroce, Michael and Shah, Shital and Shang, Ning and Sharma, Hiteshi and Shen, Yelong and Shukla, Swadheen and Song, Xia and Tanaka, Masahiro and Tupini, Andrea and Vaddamanu, Praneetha and Wang, Chunyu and Wang, Guanhua and Wang, Lijuan and Wang, Shuohang and Wang, Xin and Wang, Yu and Ward, Rachel and Wen, Wen and Witte, Philipp and Wu, Haiping and Wu, Xiaoxia and Wyatt, Michael and Xiao, Bin and Xu, Can and Xu, Jiahang and Xu, Weijian and Xue, Jilong and Yadav, Sonali and Yang, Fan and Yang, Jianwei and Yang, Yifan and Yang, Ziyi and Yu, Donghan and Yuan, Lu and Zhang, Chenruidong and Zhang, Cyril and Zhang, Jianwen and Zhang, Li Lyna and Zhang, Yi and Zhang, Yue and Zhang, Yunan and Zhou, Xiren},
    month = aug,
    year = {2024},
    note = {arXiv:2404.14219 [cs]},
}

@misc{beltagy_longformer_2020,
    title = {Longformer: {The} {Long}-{Document} {Transformer}},
    shorttitle = {Longformer},
    url = {http://arxiv.org/abs/2004.05150},
    doi = {10.48550/arXiv.2004.05150},
    urldate = {2025-11-09},
    publisher = {arXiv},
    author = {Beltagy, Iz and Peters, Matthew E. and Cohan, Arman},
    month = dec,
    year = {2020},
    note = {arXiv:2004.05150 [cs]},
}

@inproceedings{li_enhancing_2024,
    title = {Enhancing {Visual} {Document} {Understanding} with {Contrastive} {Learning} in {Large} {Visual}-{Language} {Models}},
    url = {https://openaccess.thecvf.com/content/CVPR2024/html/Li_Enhancing_Visual_Document_Understanding_with_Contrastive_Learning_in_Large_Visual-Language_CVPR_2024_paper.html},
    language = {en},
    urldate = {2025-11-11},
    author = {Li, Xin and Wu, Yunfei and Jiang, Xinghua and Guo, Zhihao and Gong, Mingming and Cao, Haoyu and Liu, Yinsong and Jiang, Deqiang and Sun, Xing},
    year = {2024},
    pages = {15546--15555},
}

@inproceedings{xu_mlevlm_2024,
    address = {Bangkok, Thailand},
    title = {{MLeVLM}: {Improve} {Multi}-level {Progressive} {Capabilities} based on {Multimodal} {Large} {Language} {Model} for {Medical} {Visual} {Question} {Answering}},
    shorttitle = {{MLeVLM}},
    url = {https://aclanthology.org/2024.findings-acl.296/},
    doi = {10.18653/v1/2024.findings-acl.296},
    urldate = {2025-11-11},
    booktitle = {Findings of the {Association} for {Computational} {Linguistics}: {ACL} 2024},
    publisher = {Association for Computational Linguistics},
    author = {Xu, Dexuan and Chen, Yanyuan and Wang, Jieyi and Huang, Yue and Wang, Hanpin and Jin, Zhi and Wang, Hongxing and Yue, Weihua and He, Jing and Li, Hang and Huang, Yu},
    editor = {Ku, Lun-Wei and Martins, Andre and Srikumar, Vivek},
    month = aug,
    year = {2024},
    pages = {4977--4997},
}

@inproceedings{chen_spatialvlm_2024,
    title = {{SpatialVLM}: {Endowing} {Vision}-{Language} {Models} with {Spatial} {Reasoning} {Capabilities}},
    shorttitle = {{SpatialVLM}},
    url = {https://openaccess.thecvf.com/content/CVPR2024/html/Chen_SpatialVLM_Endowing_Vision-Language_Models_with_Spatial_Reasoning_Capabilities_CVPR_2024_paper.html},
    language = {en},
    urldate = {2025-11-12},
    author = {Chen, Boyuan and Xu, Zhuo and Kirmani, Sean and Ichter, Brain and Sadigh, Dorsa and Guibas, Leonidas and Xia, Fei},
    year = {2024},
    pages = {14455--14465},
}

@inproceedings{sahoo_comprehensive_2024,
    address = {Miami, Florida, USA},
    title = {A {Comprehensive} {Survey} of {Hallucination} in {Large} {Language}, {Image}, {Video} and {Audio} {Foundation} {Models}},
    url = {https://aclanthology.org/2024.findings-emnlp.685/},
    doi = {10.18653/v1/2024.findings-emnlp.685},
    urldate = {2025-11-12},
    booktitle = {Findings of the {Association} for {Computational} {Linguistics}: {EMNLP} 2024},
    publisher = {Association for Computational Linguistics},
    author = {Sahoo, Pranab and Meharia, Prabhash and Ghosh, Akash and Saha, Sriparna and Jain, Vinija and Chadha, Aman},
    editor = {Al-Onaizan, Yaser and Bansal, Mohit and Chen, Yun-Nung},
    month = nov,
    year = {2024},
    pages = {11709--11724},
}

@inproceedings{sarkar_mitigating_2025,
    address = {Suzhou, China},
    title = {Mitigating {Hallucinations} in {Vision}-{Language} {Models} through {Image}-{Guided} {Head} {Suppression}},
    isbn = {979-8-89176-332-6},
    url = {https://aclanthology.org/2025.emnlp-main.631/},
    urldate = {2025-11-08},
    booktitle = {Proceedings of the 2025 {Conference} on {Empirical} {Methods} in {Natural} {Language} {Processing}},
    publisher = {Association for Computational Linguistics},
    author = {Sarkar, Sreetama and Che, Yue and Gavin, Alex and Beerel, Peter Anthony and Kundu, Souvik},
    editor = {Christodoulopoulos, Christos and Chakraborty, Tanmoy and Rose, Carolyn and Peng, Violet},
    month = nov,
    year = {2025},
    pages = {12492--12511},
}

@inproceedings{shi_thorough_2024,
    address = {Miami, Florida, USA},
    title = {A {Thorough} {Examination} of {Decoding} {Methods} in the {Era} of {LLMs}},
    url = {https://aclanthology.org/2024.emnlp-main.489/},
    doi = {10.18653/v1/2024.emnlp-main.489},
    urldate = {2025-11-13},
    booktitle = {Proceedings of the 2024 {Conference} on {Empirical} {Methods} in {Natural} {Language} {Processing}},
    publisher = {Association for Computational Linguistics},
    author = {Shi, Chufan and Yang, Haoran and Cai, Deng and Zhang, Zhisong and Wang, Yifan and Yang, Yujiu and Lam, Wai},
    editor = {Al-Onaizan, Yaser and Bansal, Mohit and Chen, Yun-Nung},
    month = nov,
    year = {2024},
    pages = {8601--8629},
}

@inproceedings{artzy_attend_2024,
    address = {Miami, Florida, US},
    title = {Attend {First}, {Consolidate} {Later}: {On} the {Importance} of {Attention} in {Different} {LLM} {Layers}},
    shorttitle = {Attend {First}, {Consolidate} {Later}},
    url = {https://aclanthology.org/2024.blackboxnlp-1.10/},
    doi = {10.18653/v1/2024.blackboxnlp-1.10},
    urldate = {2025-11-13},
    booktitle = {Proceedings of the 7th {BlackboxNLP} {Workshop}: {Analyzing} and {Interpreting} {Neural} {Networks} for {NLP}},
    publisher = {Association for Computational Linguistics},
    author = {Artzy, Amit Ben and Schwartz, Roy},
    editor = {Belinkov, Yonatan and Kim, Najoung and Jumelet, Jaap and Mohebbi, Hosein and Mueller, Aaron and Chen, Hanjie},
    month = nov,
    year = {2024},
    pages = {177--184},
}
}


\end{document}